\documentclass[10pt,journal,compsoc]{IEEEtran}
\pdfoutput=1
\ifCLASSOPTIONcompsoc
  \usepackage[nocompress]{cite}
\else
  \usepackage{cite}
\fi

\ifCLASSINFOpdf
\else
\fi
\usepackage{hyperref}
\usepackage{amsmath}
\usepackage{amsfonts}
\usepackage{amssymb}
\usepackage{graphicx}
\usepackage{caption}
\usepackage{subcaption}
\usepackage{epstopdf}
\usepackage{multirow}
\usepackage{cite}
\usepackage{enumitem}
\usepackage{makecell}

\begin{document}

\title{Convolutional Neural Networks for \\ Non-iterative Reconstruction of \\Compressively Sensed Images}
\author{Suhas~Lohit,
        Kuldeep~Kulkarni,~Ronan~Kerviche,~Pavan~Turaga~and~Amit~Ashok
\IEEEcompsocitemizethanks{\IEEEcompsocthanksitem S. Lohit, K. Kulkarni and P. Turaga were with the Department of Electrical, Computer and Electrical Engineering and the Department of Arts, Media and Engineering, Arizona State University, Tempe, AZ\protect\\
E-mail: slohit@asu.edu, kkulkar1@asu.edu, pturaga@asu.edu 
\IEEEcompsocthanksitem R. Kerviche and A. Ashok were with the College of Optical Sciences, University of Arizona, Tucson, AZ\protect\\
E-mail: rkerviche@optics.arizona.edu, ashoka@optics.arizona.edu }
}

\IEEEtitleabstractindextext{%
\begin{abstract}
Traditional algorithms for compressive sensing recovery are computationally expensive and are ineffective at low measurement rates. In this work, we propose a data driven non-iterative algorithm to overcome the shortcomings of earlier iterative algorithms. Our solution, \textbf{ReconNet}, is a deep neural network, whose parameters are learned end-to-end to map block-wise compressive measurements of the scene to the desired image blocks. Reconstruction of an image becomes a simple forward pass through the network and can be done in real-time. We show empirically that our algorithm yields reconstructions with higher PSNRs compared to iterative algorithms at low measurement rates and in presence of measurement noise. We also propose a variant of ReconNet which uses adversarial loss in order to further improve reconstruction quality. We discuss how adding a fully connected layer to the existing ReconNet architecture allows for jointly learning the measurement matrix and the reconstruction algorithm in a single network. Experiments on real data obtained from a block compressive imager show that our networks are robust to unseen sensor noise. Finally, through an experiment in object tracking, we show that even at very low measurement rates, reconstructions using our algorithm possess rich semantic content that can be used for high level inference.

\end{abstract}

}

\maketitle
\IEEEdisplaynontitleabstractindextext
\IEEEpeerreviewmaketitle

\IEEEraisesectionheading{\section{Introduction}\label{sec:introduction}}
\IEEEPARstart{I}{mages} and video data are now ubiquitous and computer vision has grown tremendously with new applications being developed continuously in health-care, defense etc. Depending on the application, many constraints may arise when we build devices and algorithms to be deployed in the real world. In this paper, we focus on two such constraints. Sensor costs can be prohibitively expensive in certain imaging modalities. For example, in short-wave infrared (SWIR) and medium-wave infrared (MWIR) imaging, the sensor cost can dominate the entire imaging system cost. Bandwidth and power constraints arise in mobile devices, surveillance applications, imagers in space probes etc. An effective way of designing algorithms, while satisfying these constraints to a large extent, is through compressive sensing. 

Compressive Sensing (CS) is a signal acquisition paradigm that integrates sampling and compression into a single step performed by front end hardware. CS theory tells us that one can acquire a small (relative to the ambient dimension) number of measurements which are random projections of a sparse signal and later reconstruct the entire signal perfectly by solving an inverse problem \cite{Candes}. In the case of natural images, sparsity or compressibility of natural images in transform domains (such as wavelets) is exploited for this purpose. This sub-Nyquist sampling feature of CS is particularly attractive in applications where sensing time (e.g. magnetic resonance imaging) or bandwidth (e.g. surveillance) is a constraint. One of the first camera architectures to be developed based on CS is the single pixel camera \cite{SPC} and is commercially produced by the InView Corporation\footnote{\url{http://inviewcorp.com/technology/compressive-sensing/}}. This camera, as the name suggests, consists of just one photodiode that operates at the required wavelength and thus is suitable in applications like SWIR imaging where the sensor cost can become prohibitive. It is also worth mentioning that effort has been made in miniaturizing compressive imagers for possible use in mobile phones and other handheld devices, cf. \cite{CMOS}. Here, the authors also show that such sensors can be more energy efficient than their traditional counterparts. 

In order to employ such a camera in computer vision for image recognition, tracking etc., a natural pipeline emerges.  Once the image is reconstructed from the low-dimensional CS measurements, existing computer vision algorithms can be used without modification. However, iterative reconstruction algorithms form a computational bottleneck in the pipeline. It may take as much as 5 minutes to reconstruct a single image of size $256 \times 256$ using one of these algorithms. This is unacceptable in applications where inference needs to be done in real-time. These algorithms are also ineffective at low measurement rates below 0.1 for images, which is where the advantages of CS are most evident in terms of data reduction. Also, the sparsity level and the sparsifying basis may need to be known by the user, which is usually an ad hoc choice. In this paper, we propose a new reconstruction algorithm that overcomes these drawbacks and is capable of yielding good quality images in real time. Inspired by the recent success of deep Convolutional Neural Networks (CNNs) in computer vision tasks such as super-resolution \cite{dong2014learning, dong2016image}, semantic segmentation \cite{girshick2014rich}, \cite{Long2015fully} etc., we design a novel architecture to map compressive measurements of an image block to the reconstructed image block. Once the architecture (and other hyper-parameters such as the learning rate schedule) is fixed, our approach is entirely data-driven which means that all parameters of the network are learned end-to-end based on training data. We now outline the main contributions of this paper.   
	
\subsection*{Contributions}
\begin{enumerate}
	
	\item We propose a novel CS reconstruction algorithm, called ReconNet, that is non-iterative and 3 orders of magnitude faster than conventional iterative approaches. Based on the loss function, we have two variants of ReconNet -- \textbf{ReconNet (Euc)} trained using Euclidean loss and \textbf{ReconNet (Euc + Adv)} trained using a combination of Euclidean and adversarial loss. 
	
	\item We carry out extensive experiments on a standard test dataset by simulating CS in software and show that our algorithm produces superior quality reconstruction in terms of PSNR at low measurement rates of 0.1 and below, as well as in the presence of measurement noise. We also compare the two variants and show that adding adversarial loss results in sharper reconstructions and improved PSNRs at higher measurement rates.

	\item We demonstrate the robustness of our network to arbitrary sensor noise by showing high quality reconstructions from real CS measurements obtained using a scalable block compressive camera, although the network is trained using a synthetic set.
	
	\item The network complexity of ReconNet is concentrated in the first fully connected layer which accounts for more than 80\% of the parameters at higher measurement rates. We propose circulant layers as an alternative to this layer which greatly reduces the number of weights. We verify experimentally that even with a 95\% reduction in the number of parameters in the first layer, the drop in PSNR is within 2 dB for a wide range of measurement rates. 
	
	\item Finally, we make the important observation that even reconstructions at very low measurement rates of about 0.01 retain sufficient semantic content that allow for effective high-level inference such as object tracking.
	
\end{enumerate}

\bigskip

\noindent \textbf{This paper is an extension of its preliminary version that appeared in CVPR 2016 \cite{Kulkarni_2016_CVPR}.} In Section \ref{sec:gan}, we modify the loss function to include adversarial loss which gives sharper reconstructions and higher PSNRs. In Section \ref{sec:learn_mm}, we describe joint learning of the measurement matrix and the reconstruction algorithm and show supporting results. In Section \ref{sec:circulant}, circulant layers are used to reduce the network complexity. In Sections \ref{sec:real_data} and \ref{sec:tracking}, additional results on reconstruction of real data and object tracking are presented based on the new variants of ReconNet proposed here.


\section{Background and Related Work} \label{sec:background}
We review relevant literature from compressive sensing, computer vision and deep learning here. 

\subsection{Compressive Sensing}
As mentioned in Section \ref{sec:introduction}, compressive sensing (CS) or compressive sampling is a relatively new paradigm in signal processing developed in the mid 2000s \cite{Candes}. Here, we have a linear signal acquisition model (performed by hardware) as follows. For a signal $\mathbf{x} \in \mathbb{R}^n$, the measurement vector obtained via CS, henceforth referred to as compressive measurements, denoted by $\mathbf{y} \in \mathbb{R}^m$ is given by 

\begin{equation} \label{eq:cs}
\mathbf{y} = \Phi\mathbf{x}, \quad m<<n,
\end{equation}

where $\Phi \in \mathbb{R}^{m \times n}$ is called the measurement matrix. Recovering $\mathbf{x}$ from $\mathbf{y}$ is an inverse problem and not admit a unique solution in general. Researchers have shown theoretically that as long as $m = O(s\log(\frac{n}{s}))$, where $s$ is the number of non-zeros in $\mathbf{x}$ when expressed in a transform domain $\Psi$ and the entries of $\Phi$ are drawn from a sub-Gaussian distribution such as a Gaussian, Bernoulli etc., it is possible to recover $\mathbf{x}$ from $\mathbf{y}$ perfectly \cite{donoho2006compressed}, \cite{candes2006near}. In this paper, the data type of interest is natural images and it is worth mentioning that natural images, although not sparse, are "compressible" in the wavelet domain. The recovery/reconstruction problem has received a great amount of attention in the past decade and we briefly discuss the main algorithms and their drawbacks next.

\subsubsection*{Iterative algorithms for reconstruction}
Several algorithms have been proposed to reconstruct images from CS measurements.  The earliest algorithms leveraged the traditional CS theory described above \cite{donoho2006compressed,candes2006near, candes2006robust} and solved the $l_1$-minimization in Eq. \ref{eq:l1} with the assumption that the image is sparse in some transform-domain like wavelet, DCT, or gradient. 

\begin{equation}\label{eq:l1}
\min_{\mathbf{x}} \quad||\mathbf{\Psi} \mathbf{x}||_1  \quad \quad s.t \quad \quad  ||\mathbf{y} - {\bf \Phi} \mathbf{x}||_2 \le \epsilon.
\end{equation}

However,  such sparsity-based algorithms did not work well, since images, though compressible, are not exactly sparse in the transform domain.  This heralded an era of model-based CS recovery methods, wherein more complex image models that go beyond simple sparsity were proposed. Model-based CS recovery methods come in two flavors.  In the first, the image model is enforced explicitly  \cite{duarte2008wavelet,  baraniuk2010model, kim2010compressed, som2012compressive}, wherein in each iteration the image estimate is projected onto the solution set defined by the model. These models, often considered under the class of `structured-sparsity' models, are capable of capturing the higher order dependencies between the wavelet coefficients. However, generally a computationally expensive optimization is solved to obtain the projection.
In the second, the algorithms enforce the image model implicitly through a non-local regularization term in the objective function \cite{peyre2008non,  zhang2013improved,  dong2014compressive}. Recently, a new class of recovery methods called approximate message passing (AMP) algorithms \cite{donoho2009message, tan2015compressive, metzler2014denoising} have been proposed, wherein the image estimate is refined in each iteration using an off-the-shelf denoiser.

\subsection{CNNs for per-pixel prediction tasks}
Computer vision, amongst other fields, has undergone a transformation since the re-introduction of convolutional neural networks (CNNs), now armed with a lot more labeled data (e.g. ImageNet), computational power (GPUs) and algorithmic improvements (ReLu, Dropout) \cite{krizhevsky2012imagenet}. CNNs have been particularly attractive and more powerful compared to their connected counterparts because CNNs are able to exploit the spatial correlations present in natural images and each convolutional layer has far less learnable parameters than a fully-connected layer and thus is less prone to overfitting. CNNs learn high-level non-linear features directly from data and have outperformed all other algorithms (and even humans in some cases) for high level inference problems like image recognition, face recognition and object detection.

In addition to inference, there has been a great amount of exciting research in areas like semantic segmentation \cite{girshick2014rich}, \cite{Long2015fully}, depth estimation \cite{eigen2014depth}, surface normal estimation \cite{Wang2015surface} etc., where CNNs have outperformed all traditional methods. In such an application, an input image is mapped to a similar-sized output. Another related class of tasks which is of interest here can be termed as inverse problems -- problems where the output is of a higher dimension than that of the input. Example include automatic colorization \cite{cheng2015deep}, 3D reconstruction a single image \cite{categoryShapesKar15} and super-resolution (SR) \cite{dong2014learning}, \cite{dong2016image}. For SR, the authors design a CNN, SRCNN, that takes an input image that is upsampled using bicubic interpolation and produces a super-resolved version of the original image of a lower resolution. The network architecture we design in this paper is inspired by SRCNN. The reason for this is that the problem of CS reconstruction can be seen as a generalization of SR. However, although both CS recovery and SR can be cast as solving an inverse problem $\mathbf{y} = \Phi\mathbf{x}$, they are not considered under the same umbrella. The reason for this is discussed in more detail in Section \ref{sec:reconnet}.

To summarize, we make the observation that any neural network can be viewed as an algorithm that allows for efficient learning of a non-linear mapping from the input to the desired output. In our case, we push this notion to the extent of learning the (necessarily non-linear) mapping from CS measurements to the image. This is also significant since 2D CNNs have until now been mainly shown to be useful for inputs which are images. CS measurements, however, are typically random projections of the scene and do not have the spatial correlational structure present in natural images. Thus, they cannot be used directly as inputs to a 2D CNN. In Section \ref{sec:reconnet}, we describe the architecture that aims at resolving this apparent incompatibility.  

\subsubsection{Generative Adversarial Networks} \label{sec:ganintro}
In section \ref{sec:gan}, we discuss a modification of the loss function for ReconNet based on the recently popular Generative Adversarial Network (GAN) framework. It has been shown recently that for inverse problems such as image inpainting \cite{pathak2016context}, super-resolution \cite{ledig2016photo} and surface normal estimation \cite{yoon2016fine}, using a GAN framework yields sharper results, than by using just Eucliden loss. This is simply due to the averaging effect of Euclidean loss minimization. As described in the papers by Goodfellow et al \cite{goodfellow2014generative} and Radford et al. \cite{radford2015unsupervised}, i.e., in the original formulation, a GAN learns to model the image distribution where the image is represented as a random variable $R$ in an unsupervised fashion by learning a mapping from a small dimensional uniform random variable, $z$ (which we can sample easily) to the image. In a GAN, two networks -- a generator, $\mathcal{G}$ with parameters $\Theta_{\mathcal{G}}$ and a discriminator, $\mathcal{D}$, with parameters  $\Theta_{\mathcal{D}}$ -- are trained in an alternating fashion. $\mathcal{G}$ is a neural network responsible for generating an image for a given input. $\mathcal{D}$ is another neural network which learns to classify between ``real" images and images output by the generator -- ``fake" images. During training, $\mathcal{D}$ tries to minimize this classification error by updating $\Theta_{\mathcal{D}}$. At the same time, $\mathcal{G}$ tries to maximize the loss of $\mathcal{D}$ by updating $\Theta_{\mathcal{G}}$, thereby trying to ``fool" $\mathcal{D}$. The mathematical form of the optimization is given by 

\begin{equation}
\underset{\Theta_{\mathcal{G}}}{\text{min}} \, \underset{\Theta_{\mathcal{D}}}{\text{max}} \quad \mathbb{E}_{R}[ \log(\mathcal{D}(r))] + \mathbb{E}_{Z} [\log(1 - \mathcal{D}(\mathcal{G}(z)))]
\end{equation}

Theoretically, it has been shown that the above optimization results in $\mathcal{D}$ being unable to classify better than chance and  $\mathcal{G}$ learning to model the data distribution and generate "realistic" images.

\subsection{Purely data-driven approaches to CS image and video reconstruction using deep learning} Ali et al. \cite{mousavi2015deep} first presented a stacked denoising auto-encoders (SDAs) based non-iterative approach for problem of CS reconstruction. In the preliminary version of this paper \cite{Kulkarni_2016_CVPR}, we proposed a convolutional architecture, which has fewer parameters, and is easily scalable to larger block-size at the sensing stage. One of the drawbacks of the approaches presented in both \cite{mousavi2015deep} and \cite{Kulkarni_2016_CVPR} is that the reconstructions are performed independently on each block. It results in the approaches not utilizing the strong dependencies that exist between the reconstructions of different blocks. In order to address this, Ali et al. \cite{mousavi2017learning} propose a network that can operate on the CS measurements of the entire image, while forcing the fully connected layer to be $\Phi^{T}$. Ali et al. propose another method which learns to simultaneously compute non-linear measurements and the reconstruction layers using an autoencoder framework \cite{mousavi2017deepcodec}. Yao et al.\cite{yao2017dr} modify the ReconNet architecture \cite{Kulkarni_2016_CVPR} by adding residual connections and present improved reconstruction performance. Dave et al. \cite{dave2016compressive} show that by enforcing an image prior which captures long term spatial dependencies, one can recover better quality reconstructions than the iterative counterparts. Chakrabarti \cite{chakrabarti2016learning}, instead of using random measurements, proposes to learn the measurement matrix in conjunction with the non-iterative reconstruction network. The success of the deep learning approaches in compressive recovery problem has not been limited to the image reconstruction problem. Researchers have shown that they can be applied to the CS video recovery problems as well \cite{iliadis2016deep, xu2016csvideonet}.


\begin{figure*}[ht!]
		\centering
		\includegraphics[trim={1cm, 5cm, 7.5cm, 19cm}, clip, height = 2 in]{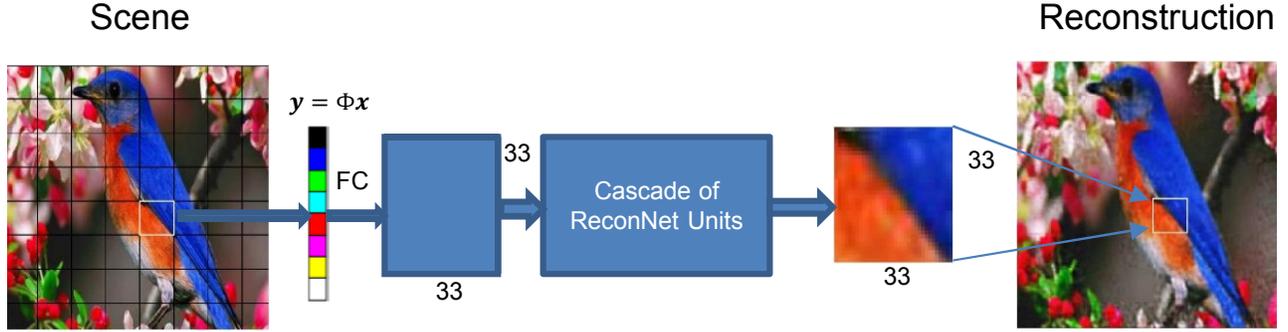}
	\caption{This shows the overview of the proposed non-iterative CS reconstruction algorithm: ReconNet. The architecture that we use for all the experiments operates using block CS. A scene is divided into blocks of size $33 \times 33$ and CS measurments of each block is passed through the ReconNet to obtain the reconstructed image patch. As a post-processing step, the image thus obtained is passed through BM3D denoiser to get rid of the blocky artifacts.}
	\label{fig:reconnet}
\end{figure*}

\section{ReconNet}\label{sec:reconnet}
In this section, we describe in detail the network architecture and other implementation details. Figure \ref{fig:reconnet} shows the overview of the proposed algorithm. Each image is divided into non-overlapping whose CS measurements are obtained separately. We need to reconstruct each image block from its compressive measurements. Then, the block reconstructions are arranged to form an image and passed through an off-the-shelf denoiser to remove the blocky artifacts and produce the final reconstruction.

Although our network architecture was inspired by SRCNN \cite{dong2014learning}, \cite{dong2016image}, the input in our case is a one-dimensional vector of CS measurements without any spatial structure, unlike an image in the case of SRCNN. Thus, in order to employ a CNN for reconstruction, we need to first resolve this incompatibility. One way to work around this problem is to seek inspiration from the SRCNN pipeline where an initial high resolution image is first obtained using bicubic interpolation and is used as the input to 3-layer CNN which produces the final super-resolved image. In our case, we could use an initial image estimate obtained using one of the many iterative approaches and then use the network to refine it to produce the final reconstruction. Although this is straightforward conceptually, the question of how many iterations of the algorithm to run to get the initial image estimate is hard to answer. While increasing the number of iterations improves the initial estimate, it also increases the run-time, thus moving away from the goal of fast implementation. On the other hand, too few iterations yield poor estimates. Therefore, we opt for a better and a more elegant solution -- to use a fully connected layer at the beginning in order map the CS measurement vector to a two-dimensional array that may serve as an initial image estimate. However, all the parameters of the network are learned end-to-end. The presence of the fully connected layer, is also the main reason why we need to operate block-wise instead of trying to reconstruct the whole image in directly. If we were to do the latter, the number of parameters in the fully connected layer would be too large to store the weights and would be very vulnerable to overfitting. We discuss alternatives to the fully connected layer later in Section \ref{sec:circulant}.

\begin{figure}[ht!]
	\centering
	\includegraphics[trim={3cm, 10cm, 9cm, 15.25cm}, clip, height = 1in]{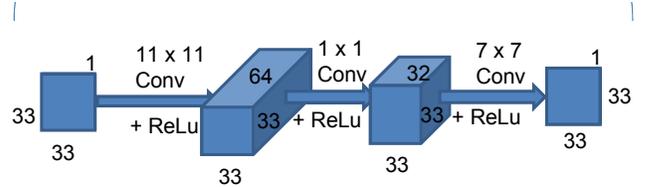}
	\caption{Each ReconNet Unit consists of 3 convolutional layers with ReLU non-linearity. Using appropriate zero-padding, the size of each feature map is always kept constant and equal to the block size.}
	\label{fig:reconnetunit}
\end{figure}

\subsection{ReconNet Unit Architecture}\label{sec:architecture}
 The input to the network is a vector of size $m \times 1$ denoted by $\Phi\mathbf{x}$, where $\Phi$ is the measurement matrix and $\mathbf{x}$ is the vectorized image block of size $n \times 1$ such that $m << n$. In all the experiments, we set the block-size to be $33 \times 33$, ($n = 1089$) since this size gives a good trade-off between reconstruction quality and network complexity. 

The first layer is a fully connected layer that takes compressive measurements as input and outputs a feature map of size $33 \times 33$. This feature map is then input to a series of `ReconNet units'. Each ReconNet unit consists of three convolutional layers as shown in Figure \ref{fig:reconnetunit}. ReLU non-linearity is employed. Using appropriate zero-padding, all feature maps produced by all convolutional layers are set to size $33 \times 33$, which is equal to the block size.

The first convolutional layer uses kernels of size $11 \times 11$ and generates 64 feature maps. The second convolutional layer uses kernels of size $1 \times 1$ and generates 32 feature maps. The third convolutional layer uses a $7 \times 7$ kernel and generates a single feature map. The output of the third layer of the last ReconNet unit is also the output of the network. 

Once all the blocks of an image have been reconstructed, the entire image is input to a denoiser to reduce blocky artifacts that arise as a result of block-wise processing. We choose BM3D \cite{dabov2007image} as the denoiser since it is fast and yields good results.

\subsection{Training Data} 
\noindent{\textbf{Ground Truth:} } We uniformly extract patches of size $33 \times 33$ from 91 natural images (these are the same images used for training in \cite{dong2014learning} and can be found on this website \footnote{\url{mmlab.ie.cuhk.edu.hk/projects/SRCNN/SRCNN_train.zip}}) with a stride equal to 14 to form a set of $21760$ patches. We retain only the luminance component of the extracted patches (during the test phase, in order to reconstruct RGB images, we replicate same network to recover the individual channels). These image blocks form the desired outputs or the ground truth of our training set. Experiments indicate that this training set is sufficient to obtain very competitive results compared to existing CS reconstruction algorithms.

\bigskip

\noindent{\textbf{Input data:}} \label{sec:training}
To train our networks, we need CS measurements corresponding to each of the extracted image blocks. To this end, we simulate noiseless CS as follows. For a given measurement rate, we construct a measurement matrix, $\Phi$ by first generating a random Gaussian matrix of appropriate size, followed by orthonormalizing its rows. Then, we apply $\mathbf{y} = \Phi\mathbf{x}$ to obtain the set of CS measurements, where $\mathbf{x}$ is the vectorized version of the luminance component of an image block. Thus, an input-label pair in the training set can be represented as ($\Phi\mathbf{x},\mathbf{x}$). We train networks for four different measurement rates (MR) = $0.25, 0.10, 0.04$ and $0.01$. Since, the total number of pixels per block is $n = 1089$, the number of measurements $n = 272, 109, 43$ and $10$ respectively. 

\subsection{Loss Function}
In this section, we describe to two variants of ReconNet based on the loss function used in training. 

\subsubsection{Euclidean Loss} \label{sec:euc} The first variant of ReconNet employs the Euclidean loss i.e., the average reconstruction error over all the training image blocks, given by

\begin{equation} \label{eq:euclidean_loss}
L(\Theta) = \frac{1}{B} \sum_{i=1}^{B} ||f(\mathbf{y_i},\Theta) - \mathbf{x}_i||^2,
\end{equation}

and is minimized by adjusting the parameters (weights and biases) in the network, $\Theta$ using mini-batch gradient descent with backpropagation. $B$ is the total number of image blocks in one batch of the training set, $x_i$ is the $i^{th}$ patch and $f(\mathbf{y_i}, \Theta)$ is the network output for $i^{th}$ patch. We set the batch size, $B = 128$ for all the networks. For each measurement rate, we train two networks, one with random Gaussian initialization for the fully connected layer, and one with a deterministic initialization, and choose the network which provides the lower loss on a validation set. For the network with deterministic initialization, the $j^{th}$ weight connecting the $i^{th}$ neuron of the fully connected layer is initialized to be equal to $\Phi^T_{i,j}$. In each case, weights of all convolutional layers are initialized using a random Gaussian with a fixed standard deviation. The learning rate is determined separately for each network using a linear search. Through experiments, we have found that two ReconNet units (6 convolutional layers in total) produce good performance. Adding further ReconNet units does not produce a significant boost in reconstruction quality and adds to network complexity. All networks are trained on an Nvidia Tesla K40 GPU using Caffe \cite{jia2014caffe} for about a day each even though the reconstruction errors converges quickly to the final value within few hours. For testing, we choose the best network by using a validation set. We refer to this network as \textbf{ReconNet (Euc)}, which uses Gaussian matrix for sensing and only the Euclidean loss function.

\subsubsection{Euclidean + Adversarial Loss} \label{sec:gan}

Here, we describe the second variant of ReconNet by incorporating the GAN framework for CS reconstruction similar to \cite{pathak2016context}. See Section \ref{sec:ganintro} for an overview of GANs and notation. In our case, \textbf{ReconNet acts as $\mathcal{G}$}. We build $\mathcal{D}$ that takes as input either the reconstructed block from ReconNet ("fake") or the desired block ("real") and outputs the probability of the input being a real image block. The loss function of $\mathcal{D}$ is the sum of two cross-entropy losses shown below:

\begin{equation}
\mathcal{L}_\mathcal{D} = \frac{1}{B} \sum_{i = 1}^{B} (\mathcal{L}_{CE} (\mathcal{D}(\mathbf{x}_i), 1) + \mathcal{L}_{CE} (\mathcal{D}(\mathcal{G}(\mathbf{y}_i)), 0)).
\end{equation}

The first loss term measures how well $\mathcal{D}$ is able to classify the real images while the second loss term measures its ability to classify the fake images generated by ReconNet, i.e, $\mathcal{G}$. Following the same notation as before, $\mathbf{y}_i$ denotes the $i^{th}$ input training CS measurement vector and $\mathbf{x}_i$ denotes the ground truth $33 \times 33$ image block associated with it.  $\mathcal{L}_{CE}()$ is the cross-entropy loss commonly used in binary classification, given by

\begin{equation}
\mathcal{L}_{CE}(\mathbf{\hat{c}},c) = -c \log{\hat{c}} + (1-c) \log{(1 - \hat{c})}
\end{equation}

The loss for $\mathcal{G}$ i.e., ReconNet is a linear combination of the Euclidean loss (from Equation \ref{eq:euclidean_loss}) and the adversarial loss:

\begin{equation} \label{eq:advloss}
\mathcal{L}_\mathcal{G} = \frac{\lambda_{rec}}{B} \sum_{i=1}^{B} ||\mathcal{G}(\mathbf{y_i}) - \mathbf{x}_i||^2 + \frac{\lambda_{adv}}{B}\sum_{i=1}^{B}\mathcal{L}_{CE} (\mathcal{D}(\mathcal{G}(\mathbf{y}_i)), 1)
\end{equation}

The protocol for initializing and training the $\mathcal{G}$  portion is the same as in the case of Euclidean loss (see Section \ref{sec:euc}). However, we use just one ReconNet unit in this case as the reconstruction quality does not improve and also becomes harder to train due to the presence of $\mathcal{D}$ in addition to $\mathcal{G}$. Since $\mathcal{G}$ is fixed, the remaining hyperparameters that need to be determined are the values of $\lambda_{rec}$, $\lambda_{adv}$ and the structure of $\mathcal{D}$, which is another, much smaller, CNN. These hyperparameters were determined by measuring the reconstruction performance on the validation set for different settings. We use a  $\mathcal{D}$ with the following architecture. It has 3 convolutional layers and each layer generates four feature maps of size $4 \times 4$ filters. At the end of the third convolutional layer, a fully connected layer maps the feature maps to a single probability value. Dropout with probability equal to 0.5 is used for this layer. $\lambda_{rec}$ and $\lambda_{adv}$ are set to 1 and 0.0001 respectively. Adam optimizer is used for learning \cite{kingma2015adam}. The learning rates for $\mathcal{G}$ and $\mathcal{D}$ are set to $10^{-3}$ and $10^{-5}$ respectively and the momentum is set to 0.9. The training of these networks is done in an alternating fashion using TensorFlow \cite{tensorflow2015-whitepaper}. We update $\Theta_{\mathcal{G}}$ twice for every update of $\Theta_{\mathcal{D}}$ since this leads to faster convergence. Training is carried out for $10^{5}$ iterations which means that $\Theta_{\mathcal{G}}$ are updated $2\times10^{5}$ times and $\Theta_{\mathcal{D}}$ are updated $10^{5}$ times. We refer to this network as \textbf{ReconNet (Euc + Adv)}, which uses Gaussian matrix for sensing and the Euclidean + adversarial loss function. 


\section{Synthetic Experiments}
In this section, we conduct extensive experiments on simulated CS data, and compare the performance of ReconNet with state-of-the-art CS image recovery algorithms, both in terms of reconstruction quality and time complexity. 

\bigskip 
\noindent {{\bf Baselines}} We compare both variants of our algorithm described in Section \ref{sec:reconnet} with three iterative CS image reconstruction algorithms,  TVAL3 \cite{li2013efficient}, NLR-CS \cite{dong2014compressive} and D-AMP \cite{metzler2014denoising}. We use the code made available by the respective authors on their websites. Parameters for these algorithms, including the number of iterations, are set to the default values. Since the reconstruction is performed block-wise, blocky artifacts arise. We use BM3D \cite{dabov2007image} denoiser to reduce these artifacts since it gives a good trade-off between time complexity and reconstruction quality. The code for NLR-CS provided on author's website is implemented only for random Fourier sampling. The algorithm first computes an initial estimate using a DCT or wavelet based CS recovery algorithm, and then solves an optimization problem to get the final estimate. Hence, obtaining a good estimate is critical to the success of the algorithm. However, using the code provided on the author's website, we failed to initialize the reconstruction for random Gaussian measurement matrix. Similar observation was reported by \cite{metzler2014denoising}. Following the procedure outlined in \cite{metzler2014denoising}, the initial image estimate for NLR-CS is obtained by running D-AMP (with BM3D denoiser) for $8$ iterations. Once the initial estimate is obtained, we use the default parameters and obtain the final NLR-CS reconstruction. 

We also compare with \cite{mousavi2015deep} which presents an SDA based non-iterative approach to recover from block-wise CS measurements. Here, we compare our algorithm with our own implementation of SDA, and show that our algorithm outperforms SDA. 

For fair comparison, we denoise the image estimates recovered by baselines as well. The only parameter to be input to the BM3D algorithm is the estimate of the standard Gaussian noise, $\sigma$. To estimate $\sigma$, we first compute the estimates of the standard Gaussian noise for each block in the intermediate reconstruction given by $\sigma_i = \sqrt{\frac{||y_i - \Phi x_i||^2}{m}}$, and then take the median of these estimates. 

\subsection{Reconstruction of simulated CS data}
\begin{figure*}[h]
	\centering
	\begin{subfigure}[]{0.15\textwidth}
		\centering
		\includegraphics[height=1.2in]{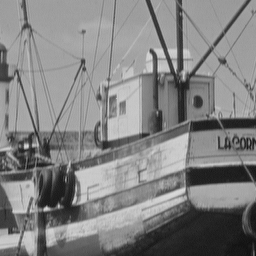}
		\caption{Boats}
	\end{subfigure}%
	\hfill
	\begin{subfigure}[]{0.15\textwidth}
		\centering
		\includegraphics[height=1.2in]{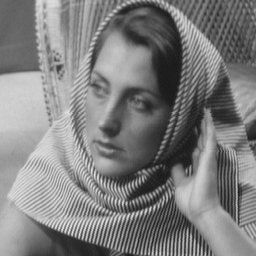}
		\caption{Barbara}
	\end{subfigure}%
	\hfill
	\begin{subfigure}[]{0.15\textwidth}
		\centering
		\includegraphics[height=1.2in]{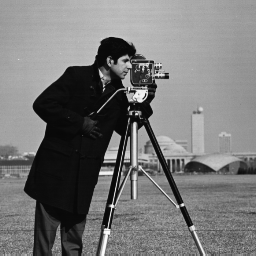}
		\caption{Cameraman}
	\end{subfigure}%
	\hfill
	\begin{subfigure}[]{0.15\textwidth}
		\centering
		\includegraphics[height=1.2in]{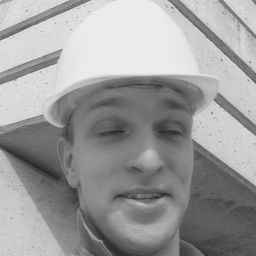}
		\caption{Foreman}
	\end{subfigure}%
	\hfill
	\begin{subfigure}[]{0.15\textwidth}
		\centering
		\includegraphics[height=1.2in]{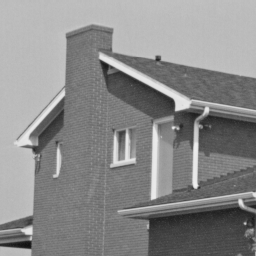}
		\caption{House}
	\end{subfigure}%
	\hfill
	\begin{subfigure}[]{0.15\textwidth}
		\centering
		\includegraphics[height=1.2in]{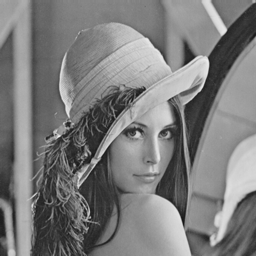}
		\caption{Lena}
	\end{subfigure}%
	
	\hfill
	\begin{subfigure}[]{0.18\textwidth}
		\centering
		\includegraphics[height=1.2in]{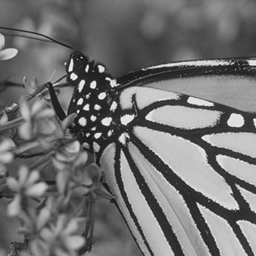}
		\caption{Monarch}
	\end{subfigure}
	\hfill
	\begin{subfigure}[]{0.18\textwidth}
		\centering
		\includegraphics[height=1.2in]{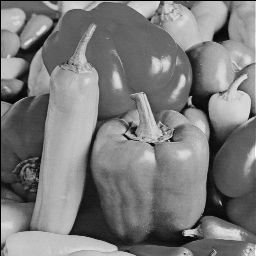}
		\caption{Peppers}
	\end{subfigure}
	\hfill
	\begin{subfigure}[]{0.18\textwidth}
		\centering
		\includegraphics[height=1.2in]{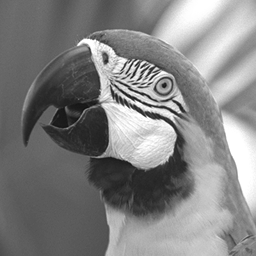}
		\caption{Parrots}
	\end{subfigure}
	\hfill
	\begin{subfigure}[]{0.18\textwidth}
		\centering
		\includegraphics[height=1.2in]{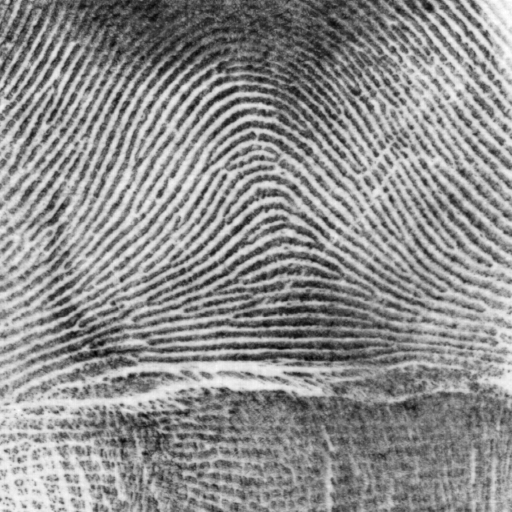}
		\caption{Fingerprint}
	\end{subfigure}
	\hfill
	\begin{subfigure}[]{0.18\textwidth}
		\centering
		\includegraphics[height=1.2in]{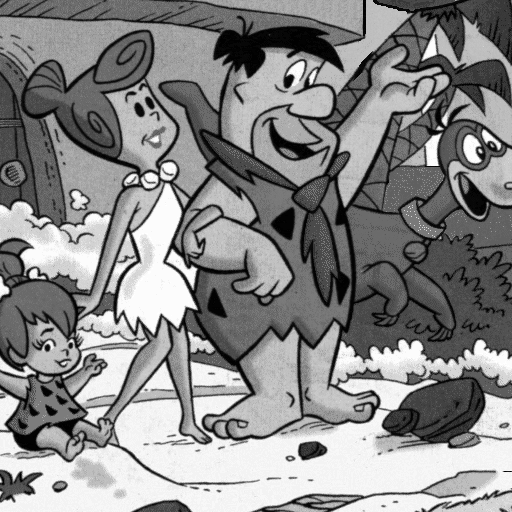}
		\caption{Flintstones}
	\end{subfigure}%
	
	\caption{Standard test set images. Note that all images are of size $256 \times 256$ (~64 non-overlapping $33 \times 33$ blocks) except Fingerprint and Flintstones which are $512 \times 512$ (~256 non-overlapping $33 \times 33$ blocks)} 
	\label{fig:test_images}
\end{figure*}

For our simulated experiments, we use a standard test set of $11$ grayscale images, compiled from two sources \footnote{\url{https://web.archive.org/web/20160403234531/http://dsp.rice.edu/software/DAMP-toolbox}} \footnote{\url{http://see.xidian.edu.cn/faculty/wsdong/NLR\textunderscore Exps.htm}}. Figure \ref{fig:test_images} shows the test images. We conduct both noiseless and noisy block-CS image reconstruction experiments at four different measurement rates 0.25, 0.1, 0.04 and 0.01.  We train two sets of networks -- The first set of networks is ReconNet Variant 1 trained with just Euclidean loss. The second set is ReconNet variant 2 trained with Euclidean + adversarial loss.

\begin{table*}
	\footnotesize
	\centering
	\begin{tabular}{|c|c|c|c|c|c|c|c|c|c|}
		\hline
		\multirow{2}{*}{\makecell{Image \\ Name}} & 
		\multirow{2}{*}{Algorithm} & 
		\multicolumn{2}{c|}{MR = 0.25} & 
		\multicolumn{2}{c|}{MR = 0.10} & 
		\multicolumn{2}{c|}{MR = 0.04} & 
		\multicolumn{2}{c|}{MR = 0.01} \\
		\cline{3-10}
		& & w/o BM3D & w/ BM3D & w/o BM3D & w/ BM3D & w/o BM3D & w/ BM3D & w/o BM3D & w/ BM3D\\
		\hline
 		
 		\multirow{5}{*}{Monarch} & TVAL3 \cite{li2013efficient} & \textbf{27.77} & \textbf{27.77} & 21.16 & 21.16 & 16.73 & 16.73 & 11.09 & 11.11\\
 		& NLR-CS \cite{dong2014compressive} & 25.91 & 26.06 & 14.59 & 14.67 & 11.62 & 11.97 & 6.38 & 6.71\\
 		& D-AMP \cite{metzler2014denoising} & 26.39 & 26.55 & 19.00 & 19.00 & 14.57 & 14.57 & 6.20 & 6.20\\
 		& SDA \cite{mousavi2015deep} & 23.54 & 23.32 & 20.95 & 21.04 & 18.09 & 18.19 & 15.31 & 15.38\\
 		& \makecell{\textbf{ReconNet} \\ (Euc)  \cite{Kulkarni_2016_CVPR}} & 24.31 & 25.06 & 21.10 & 21.51 & \textbf{18.19} & \textbf{18.32} & \textbf{15.39} & \textbf{15.49}\\
 		& \makecell{\textbf{ReconNet} \\ (Euc + Adv)} & 25.83 & 25.16 & \textbf{21.74} & \textbf{21.94} & 17.81 & 18.05 & 13.99 & 14.14 \\

 		\hline

		\multirow{5}{*}{Fingerprint} & TVAL3 & 22.70 & 22.71 & 18.69 & 18.70 & 16.04 & 16.05 & 10.35 & 10.37\\
		& NLR-CS & 23.52 & 23.52 & 12.81 & 12.83 & 9.66 & 10.10 & 4.85 & 5.18\\
		& D-AMP & 25.17 & 23.87 & 17.15 & 16.88 & 13.82 & 14.00 & 4.66 & 4.73\\
		& SDA & 24.28 & 23.45 & 20.29 & 20.31 & 16.87 & 16.83 & \textbf{14.83} & 14.82\\
		& \makecell{\textbf{ReconNet} \\ (Euc)} & 25.57 & \textbf{25.13} & 20.75 & 20.97 & 16.91 & \textbf{16.96} & 14.82 & 14.88\\
		& \makecell{\textbf{ReconNet} \\ (Euc + Adv)} & \textbf{26.19} & 24.49 & \textbf{21.21} & \textbf{21.08} & \textbf{16.97} & 16.67 & 14.78 & \textbf{14.89} \\
		
		\hline
		
		\multirow{5}{*}{Flintstones} & TVAL3 & 24.05 & 24.07 & 18.88 & 18.92 & 14.88 & 14.91 & 9.75 & 9.77\\
		& NLR-CS & 22.43 & 22.56 & 12.18 & 12.21 & 8.96 & 9.29 & 4.45 & 4.77\\
		& D-AMP & \textbf{25.02} & \textbf{24.45} & 16.94 & 16.82 & 12.93 & 13.09 & 4.33 & 4.34\\
		& SDA & 20.88 & 20.21 & 18.40 & 18.21 & 16.19 & 16.18 & 13.90 & 13.95\\
		& \makecell{\textbf{ReconNet} \\ (Euc)} & 22.45 & 22.59 & 18.92 & 19.18 & 16.30 & 16.56 & \textbf{13.96} & \textbf{14.08}\\
		& \makecell{\textbf{ReconNet} \\ (Euc + Adv)} & 24.98 & 24.38 & \textbf{20.57} & \textbf{20.36} & \textbf{16.71} & \textbf{16.85} & 13.84 & 14.02 \\
		
		\hline
		
 		\multirow{5}{*}{House} & TVAL3 & 32.08 & 32.13 & 26.29 & 26.32 & 20.94 & 20.96 & 11.86 & 11.90\\
 		& NLR-CS & \textbf{34.19} & \textbf{34.19} & 14.77 & 14.80 & 10.66 & 11.09 & 4.96 & 5.29\\
 		& D-AMP & 33.64 & 32.68 & 24.84 & 24.71 & 16.91 & 17.37 & 5.00 & 5.02\\
 		& SDA & 27.65 & 27.86 & 25.40 & 26.07 & 22.51 & 22.94 & \textbf{19.45} & \textbf{19.59}\\
 		& \makecell{\textbf{ReconNet} \\ (Euc)}& 28.46 & 29.19 & \textbf{26.69} & 26.66 & \textbf{22.58} & \textbf{23.18} & 19.31 & 19.52\\
 		& \makecell{\textbf{ReconNet} \\ (Euc + Adv)} & 30.28 & 30.92 & 26.37 & \textbf{27.19} & 22.00 & 22.58 & 18.93 & 19.17 \\

		\hline
		
		\multirow{5}{*}{\makecell{\textbf{Mean} \\ \textbf{ PSNR}}} & TVAL3 & 27.84 & 27.87 & 22.84 & 22.86 & 18.39 & 18.40 & 11.31 & 11.34\\  
		& NLR-CS & 28.05 & \textbf{28.19} & 14.19 & 14.22 & 10.58 & 10.98 & 5.30 & 5.62\\
		& D-AMP & \textbf{28.17} & 27.67 & 21.14 & 21.09 & 15.49 & 15.67 & 5.19 & 5.23\\
		& SDA & 24.72 & 24.55 & 22.43 & 22.68 & 19.96 & 20.21 & \textbf{17.29} & 17.40\\
		& \makecell{\textbf{ReconNet} \\ (Euc)} & 25.54 & 25.92 & 22.68 & 23.23 & \textbf{19.99} & \textbf{20.44} & 17.27 & \textbf{17.55}\\
		& \makecell{\textbf{ReconNet} \\ (Euc + Adv)} & 27.11 & 26.90 & \textbf{23.22} & \textbf{23.48} & 19.65 & 20.00 & 16.66 & 16.90 \\
		
		\hline
	\end{tabular}
	\vspace{0.1in}
	\caption{{PSNR values in dB for four test images as well as the mean PSNR values for the entire test set using different algorithms at different measurement rates. At low measurement rates of 0.1, 0.04 and 0.01, both variants of our algorithm yields superior quality reconstructions than the traditional iterative CS reconstruction algorithms, TVAL3, NLR-CS, and D-AMP. It is evident that the reconstructions are very stable for our algorithm with a decrease in mean PSNR of only 8.37 dB as the measurement rate decreases from 0.25 to 0.01, while the smallest corresponding dip in mean PSNR for classical reconstruction algorithms is in the case of TVAL3, which is equal to 16.53 dB. The supplement contains additional results.}}
	\label{table:psnr_test}
\end{table*}

\subsubsection*{Reconstruction from noiseless CS measurements} 
\begin{figure*}[ht!]
	\centering
	\includegraphics[width=1\textwidth]{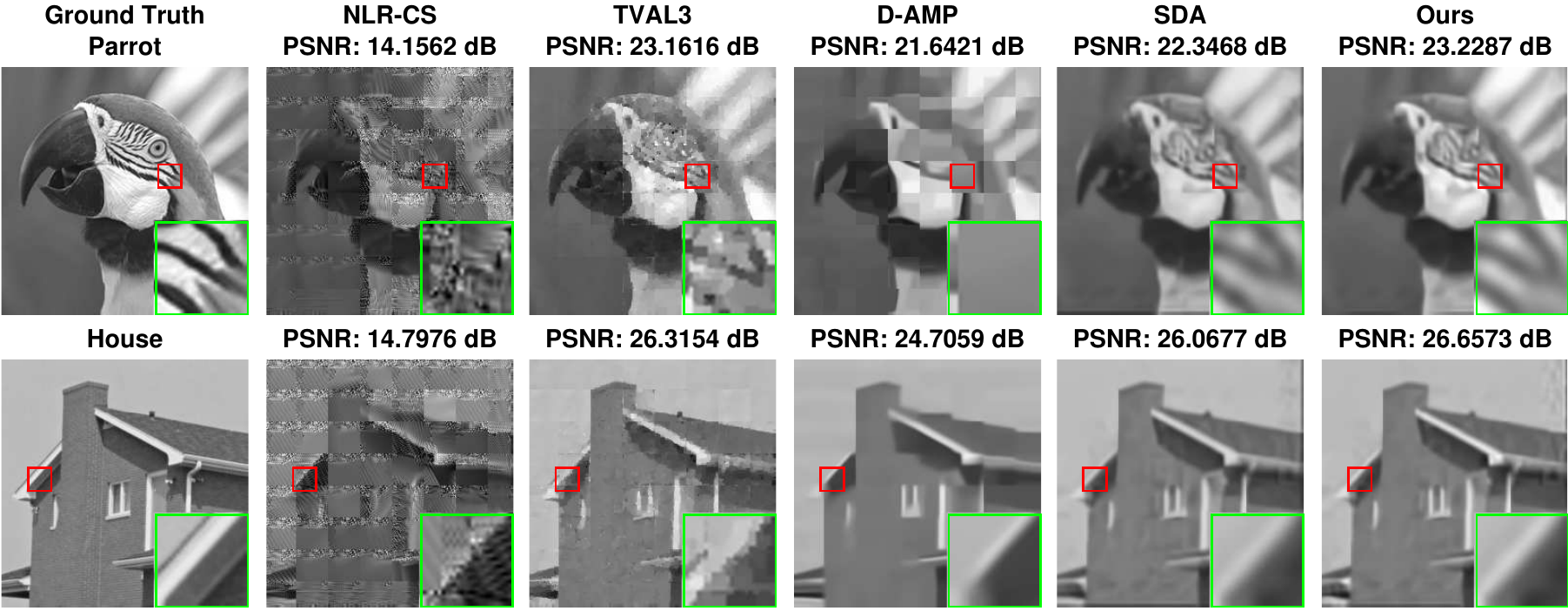}
	\caption{Comparison of reconstruction performance of various algorithms in terms of PSNR (in dB) and visual quality at MR = 0.1 and no noise for Parrot and House images. Clearly, our algorithm outperforms all the iterative algorithms. SDA also yields competetive results. The zoomed in portions show that finer structures are better retained in our case.}
	\label{fig:recon_10_sigma_0}
\end{figure*}

For a given test image, to simulate noiseless block-wise CS, we first divide the image into non-overlapping blocks of size $33 \times 33$, and then compute CS measurements for each block using Equation \ref{eq:cs}. For each measurement rate, the sensing matrix used is the same random Gaussian measurement matrix as was used to generate the training data for the network corresponding to this measurement rate in Section \ref{sec:training}. The PSNR values in dB for both reconstructions before passing through the denoiser (indicated by w/o BM3D) as well as final denoised versions (indicated by w/ BM3D) for all the measurement rates are presented in Table \ref{table:psnr_test}. It is clear from the PSNR values that both variants of our algorithm outperforms traditional reconstruction algorithms at low measurement rates of $0.1, 0.04$ and $0.01$. Also, the degradation in performance with lower measurement rates is more graceful. 
Further, in Figure \ref{fig:recon_10_sigma_0}, we show the final reconstructions of parrot and house images for various algorithms at measurement rate of 0.1 compared to ReconNet (Euc). From the reconstructed images, one can notice that our algorithm, as well as SDA, are able to retain the finer features of the images while other algorithms fail to do so. NLR-CS and DAMP provide poor quality reconstruction. Even though TVAL3 yields PSNR values comparable to our algorithm, it introduces undesirable artifacts in the reconstructions. 

 For visual comparison between ReconNet (Euc) and ReconNet (Euc + Adv), see first and second columns of Figure \ref{fig:learn_vs_gaussian}. We observe that at higher measurement rates of 0.25 and 0.10, there is improvement in reconstruction of the test set with ReconNet (Euc + Adv) over ReconNet (Euc) both in terms of PSNR ($\sim$1 dB increase) and visual quality. The reconstructed blocks are sharper than those obtained in the case of Euclidean loss. At lower measurement rates of 0.04 and 0.01, the PSNR values decrease for ReconNet (Euc + Adv) when compared to ReconNet (Euc). However, we can observe that more detail is preserved and the reconstructed images tend to be sharper when adversarial loss is used, in all cases.

\subsubsection*{Performance in the presence of noise:} 
\begin{figure}[ht!]
	\centering
	\includegraphics[width=0.48\textwidth]{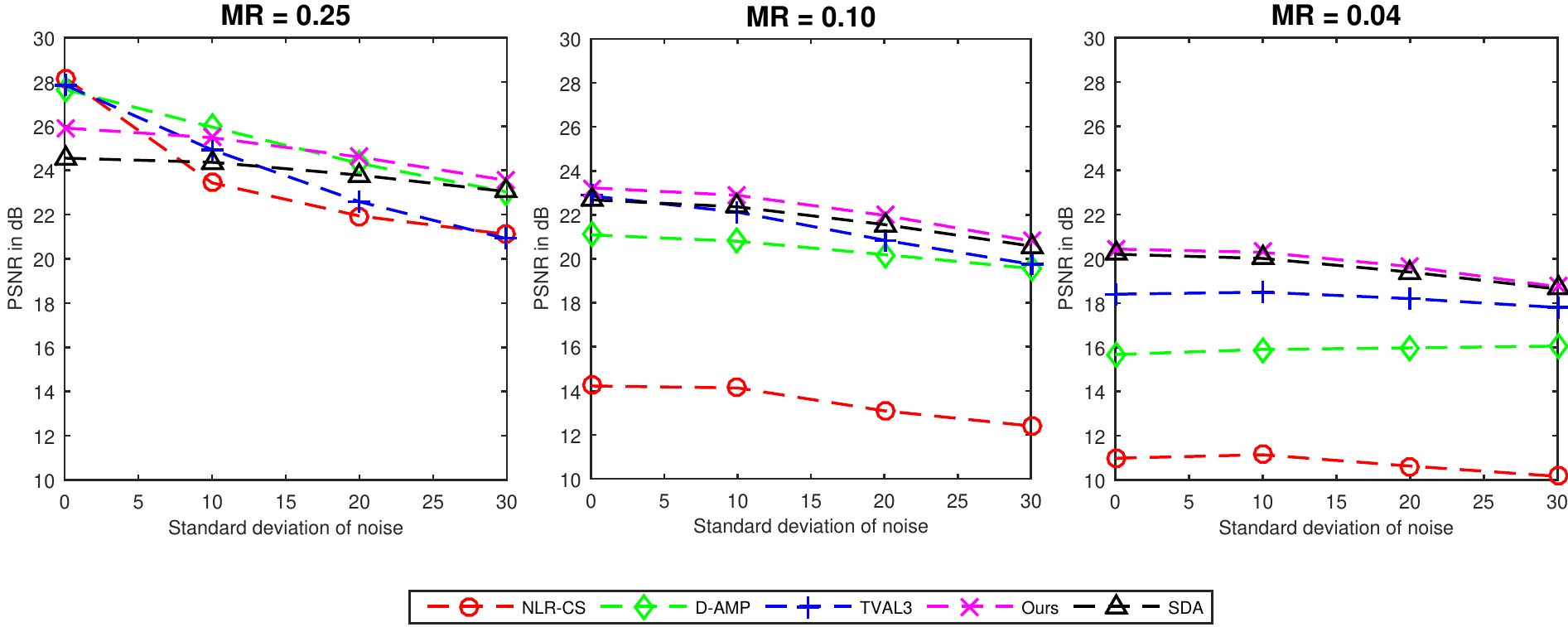}
	\caption{The figure shows the degradation of performance of various algorithms as the measurement rate and noise level in the CS measurements are increased. Our algorithm exhibits a more graceful trend compared to the iterative apporaches and outperforms them at low MRs and high noise levels.}
	\label{fig:psnr_vs_noise}
\end{figure}

We demonstrate that our algorithm is robust to Gaussian noise by performing reconstruction from noisy CS measurements. We use ReconNet (Euc) for all the experiments here and we expect the same trends to follow for other variants as well. We perform this experiment at three measurement rates - $0.25, 0.10$ and $0.04$. We emphasize that we {\bf do not} train separate networks for different noise levels but use the same networks as used in the noiseless case. In order to simulate the noisy CS process, we add standard random Gaussian noise of increasing standard deviation to the noiseless CS measurements (from the previous section) of each block. In each measurement rate, we test the algorithms at three levels of noise corresponding to $\sigma = 10,20,30$ (3.9\%, 7.8\% and 11.7\% of the dynamic range (0-255) respectively), where $\sigma$ is the standard deviation of the Gaussian noise distribution. The reconstructions obtained from the algorithms are denoised using BM3D. The mean PSNR for various noise levels for different algorithms at different measurement rates are shown in Figure \ref{fig:psnr_vs_noise}. It can be observed that our algorithm beats all other algorithms at high noise levels. This shows that the method proposed in this paper is extremely robust to all levels of noise. 

\subsection{Gains in Time Complexity}
 In addition to competitive reconstruction quality, for our algorithm without the BM3D denoiser, the computation is real-time and is about {\bf 3} orders of magnitude faster than traditional reconstruction algorithms. To this end, we compare various algorithms in terms of the time taken to produce the reconstructions of a $256 \times 256$ image from noiseless CS measurements at various measurement rates. For traditional CS algorithms, we use an Intel Xeon E5-1650 CPU to run the implementations provided by the respective authors. For ReconNet, we report computational time for both the CPU implementation of Caffe on Intel Xeon E5-1650 as well as the GPU implementation on an inexpensive mid-range Nvidia GTX 980 GPU. Note that, for our algorithm, we use a network with two ReconNet units. The average time taken for the all algorithms of interest are given in table \ref{table:time}. Depending on the measurement rate, the time taken for block-wise reconstruction of a $256 \times 256$ on the GPU for our algorithm is about $145$ to $390$ times faster than TVAL3, $1400$ to $2700$ times faster than D-AMP, and $14782$ to $15660$ times faster than NLR-CS. In the case of CPU implmenation, the speed-ups are $5.6$ to $15$ times faster, $52.9$ to $105.2$ times faster and $569$ to $600$ times faster compared to TVAL3, D-AMP and NLR-CS respectively. It is important to note that the speedup achieved by our algorithm is not solely because of the utilization of the GPU. It is because unlike traditional CS algorithms, our algorithm being CNN based relies on much simpler convolution operations, for which very fast implementations exist. More importantly, the non-iterative nature of our algorithm makes it amenable to parallelization. SDA, also a deep-learning based non-iterative algorithm shows significant speedups over traditional algorithms at all measurement rates.

\begin{table}[ht]
	\footnotesize
	\centering
	\begin{tabular}{|c|c|c|c|c|c|}
		\hline
		{Algorithm} & {MR=0.25} & {MR=0.10} & {MR=0.04} & {MR=0.01} \\
		\hline
		TVAL3 & 2.943 & 3.223 & 3.467 & 7.790\\
		\hline
		NLR-CS & 314.852 & 305.703 & 300.666 & 314.176 \\
		\hline
		D-AMP & 27.764 & 31.849 & 34.207 & 54.643\\
		\hline
		ReconNet (CPU) & 0.5249 & 0.5258 & 0.5284 & 0.5193 \\
		\hline
		ReconNet (GPU) & 0.0213 & 0.0195 & 0.0192 & 0.0244\\
		\hline
		SDA (GPU) & 0.0042 & 0.0029 & 0.0025 &  0.0045\\ 
		\hline		
		
	\end{tabular}
	\caption{Time complexity (in seconds) of various algorithms (without BM3D) for reconstructing a single $256 \times 256$ image. By taking only about 0.02 seconds at any given measurement rate, ReconNet can recover images from CS measurements in real-time, and is  {\bf $3$} orders of magnitude faster than traditional reconstruction algorithms.} 
	\label{table:time}
\end{table}
\section{Efficient Training Strategy for New Measurement Matrix}
In Section \ref{sec:training}, a new network was trained from scratch for each MR. However, it may not be practical to train a entirely new network just to operate a slightly different MR or with a different $\Phi$ at the same MR. In this section, we show that for a new $\Phi$ of a desired measurement rate, one {\bf does not} need to train the network from scratch, and that it may be sufficient to follow a suboptimal, yet effective and computationally light training strategy outlined below, ideally suited to practical scenarios.

We adapt the convolutional layers (C1-C6) of a pre-trained network for the same or slightly higher MR, henceforth referred to as the {\em base network}, and train {\bf only} the first fully connected (FC) layer with random initialization for $1000$ iterations (or equivalent time of around $2$ seconds on a Titan X GPU), while keeping C1-C6 {\bf fixed}. The network used here is ReconNet (Euc) and we expect similar trends for other variants. The mean PSNR (without BM3D) for the test set at various MRs, the time taken to train models and the MR of the base network are given in Table \ref{table:PSNR}. 
\begin{table}[ht]
	\footnotesize
	\centering
	\begin{tabular}{|c|c|c|c|c|c|}
		\hline
		\textbf{New $\Phi$ MR} & \textbf{0.1} & \textbf{0.08} & \textbf{ 0.04} & \textbf{0.01} \\
		\hline
		Base network MR & 0.25 & 0.1 & 0.1 & 0.25\\
		\hline
		Mean PSNR (dB)  & 21.73 & 20.99 & 19.66 & 16.60\\
		\hline
		Training Time (seconds) & 2 & 2 & 2 & 2\\
		\hline
	\end{tabular}
	\caption{Networks for a new $\Phi$ can be obtained by training only the FC layer of the base network at minimal computational overhead, while maintaining comparable PSNRs.}
	\label{table:PSNR}
\end{table}
From the table, it is clear that the overhead in computation for new $\Phi$ is trivial, while the mean PSNR values are comparable to the ones presented in Table \ref{table:psnr_test}. One can obtain better quality reconstructions at the cost of more training time if C1-C6 layers are also fine-tuned along with FC layer.


\section{Learning the Measurement Matrix}\label{sec:learn_mm}
Until now we have considered CS reconstruction where the measurements are acquired with a predefined sensing matrix -- a random Gaussian matrix. However, with a small addition to the the ReconNet framework, we show that it is possible to jointly, in a single network, learn the measurement matrix ($\Phi$) as well as the reconstruction algorithm. The earlier framework describes a network that map the input CS measurements to the output image block. Here, we attach an additional fully connected layer in the front that maps an input image of size $33 \times 33$ to a vector of dimension $m$. Thus the input-desired output pair in the training set is $(\mathbf{x},\mathbf{x})$. This can be seen as a variation of the autoencoder, with the constraint in the architecture that the "encoder" part of the network must be a single linear layer. This constraint arises because of the nature of the single pixel camera which can only capture linear projections of the scene. After training, the weights of the first fully connected layer correspond to the (locally) optimal measurement matrix, and the all the following layers form the reconstruction network. 

\begin{figure}
	\centering
	\begin{subfigure}[]{0.45\textwidth}
		\includegraphics[trim = {3cm, 3cm, 5cm, 1cm}, clip, width=\textwidth]{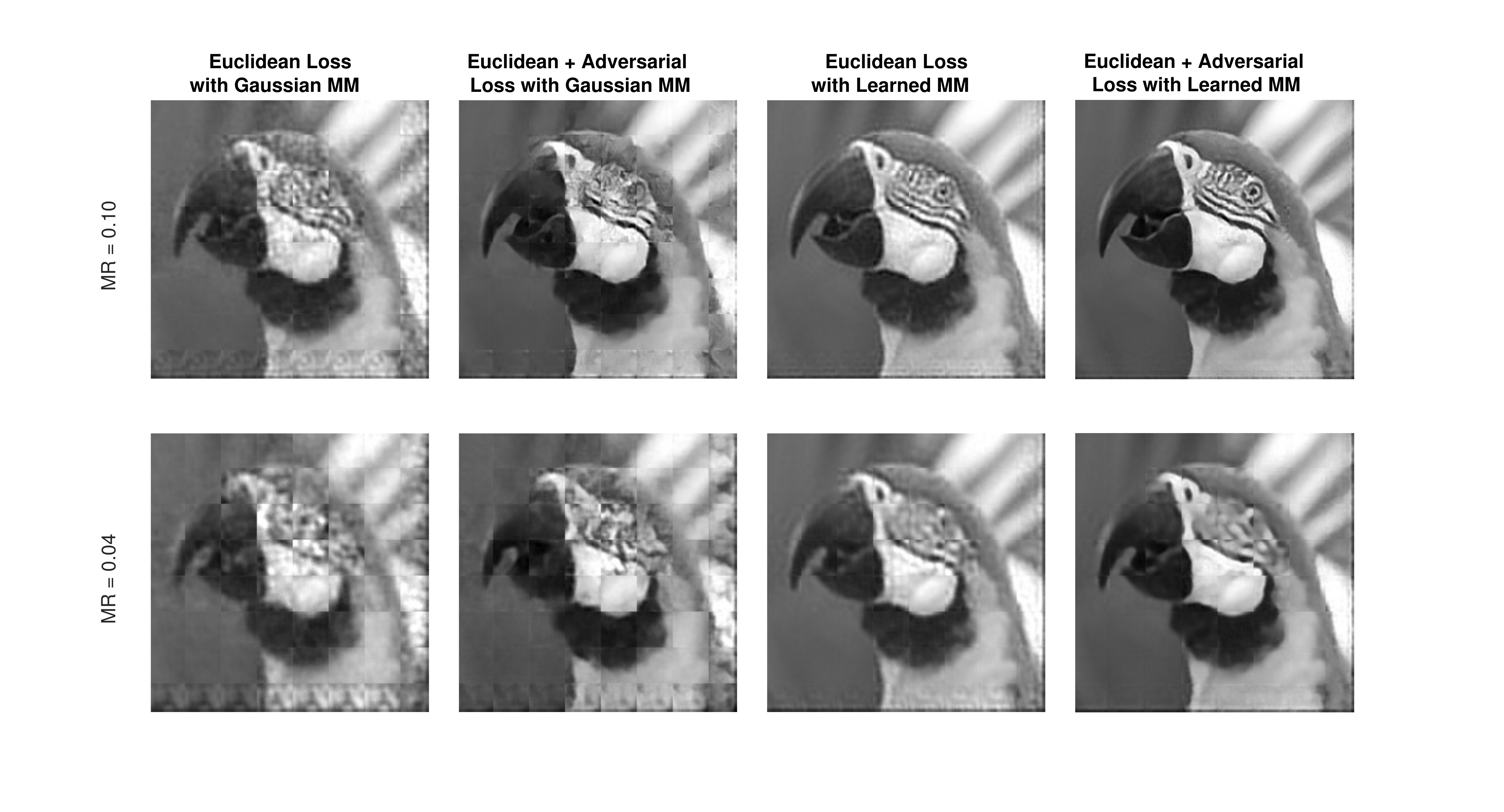}
	\end{subfigure}
	
	\begin{subfigure}[]{0.45\textwidth}
		\includegraphics[trim = {3cm, 3cm, 5cm, 1cm}, clip, width=\textwidth]{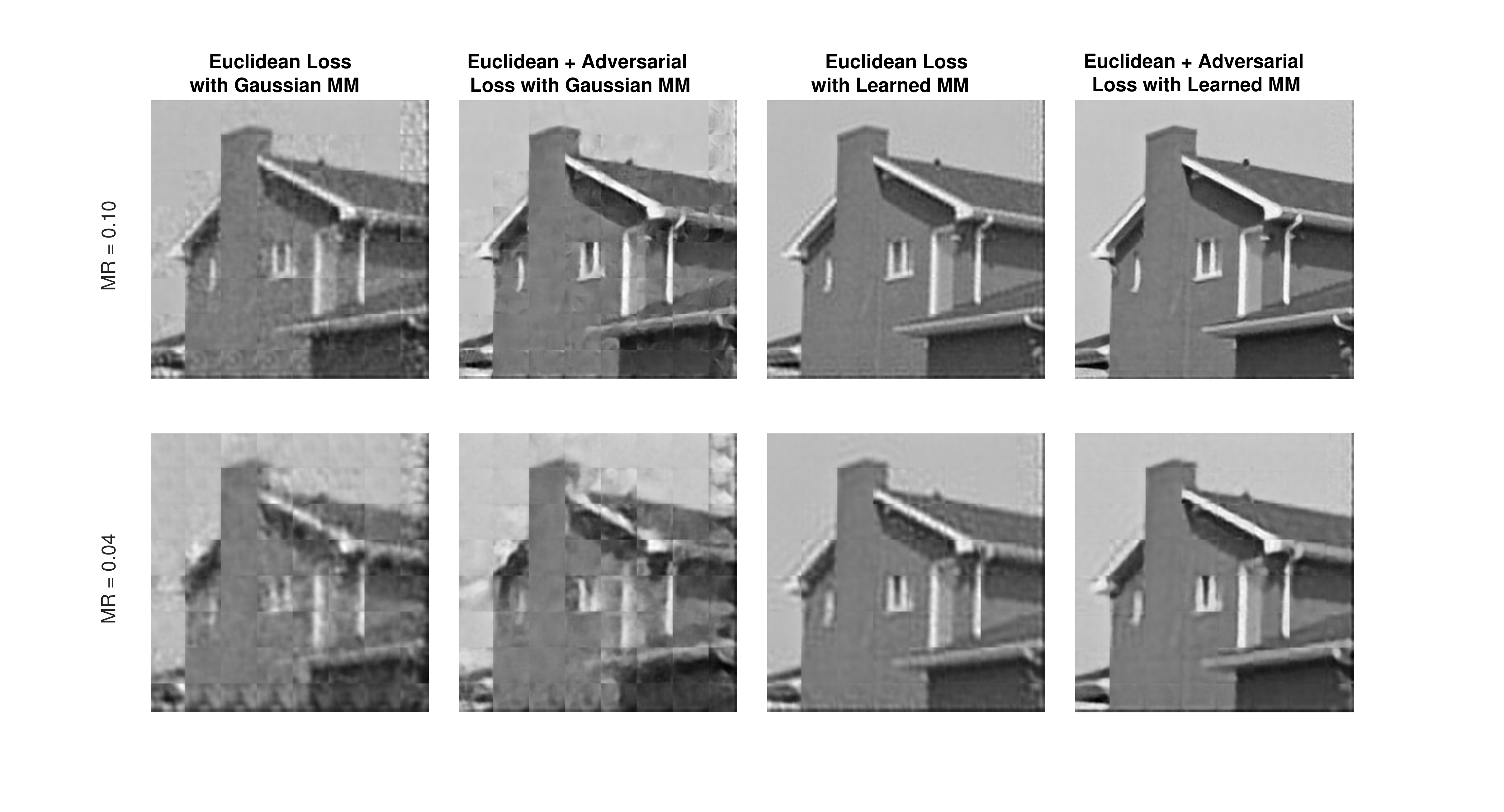}
	\end{subfigure}
	
	\caption{The figures show reconstruction results for the ``Parrot" and ``house" images at two measurement rates of 0.1 and 0.04 from measurements obtained using different variants of ReconNet. We can clearly observe that learning the measurement matrix as well as using adversarial loss while training produce superior quality reconstruction (both independently and together) at both measurement rates when compared to the basic version of ReconNet. MM refers to the measurement matrix.}
	\label{fig:learn_vs_gaussian}
\end{figure}

As before, we train two sets of networks: \textbf{ReconNet (Euc, learn $\Phi$)} -- jointly learning $\Phi$ and reconstruction algorithm using only Euclidean loss and \textbf{ReconNet (Euc + Adv, learn $\Phi$)} -- jointly learning $\Phi$ and reconstruction algorithm using Euclidean + adversarial loss. The training set consists of 21760 image patches from the same set of 91 images. Since we are learning $\Phi$ as well as the reconstruction network, each image patch in the training set forms both the input and the desired output image patch. Table \ref{table:gan} shows the mean PSNR obtained on the test set using variants of ReconNet with the learned $\Phi$ compared to ReconNet with the random Gaussian $\Phi$. We observe a significant gain in terms of PSNR at the lower measurement rates -- 2.83 dB, 3.15 dB and 2.17 dB at MR = 0.10, 0.04 and 0.01 respectively without using adversarial loss. With adversarial loss, the gains are 3.25 dB, 3.33 dB and 2.4 dB at MR = 0.10, 0.04 and 0.01 respectively. Figure \ref{fig:learn_vs_gaussian} illustrates the differences in visual quality obtained for the Parrot and House images at two different measurement rates of 0.1 and 0.04 for all four variants of ReconNet. Clearly, more detail is preserved in the case of learned $\Phi$ and using adversarial loss further sharpens reconstructions.


\begin{table*}[ht]
	\footnotesize
	\centering
	\begin{tabular}{|c|c|c|c|c|c|c|c|c|}
		\hline
		\multirow{2}{*}{\makecell{Loss function and \\ Measurement Matrix Type}} &
		\multicolumn{2}{c|}{MR = 0.25} & 
		\multicolumn{2}{c|}{MR = 0.10} & 
		\multicolumn{2}{c|}{MR = 0.04} & 
		\multicolumn{2}{c|}{MR = 0.01} \\
		\cline{2-9}
		& w/o BM3D & w/ BM3D & w/o BM3D & w/ BM3D & w/o BM3D & w/ BM3D & w/o BM3D & w/ BM3D\\
		\hline
		\makecell{Euclidean \\ with Gaussian $\Phi$}  & 25.54 & 25.92 & 22.68 & 23.23 & 19.99 & 20.44 & 17.27 & 17.55\\
		\hline
	
	\makecell{Euclidean + Adversarial \\ with Gaussian $\Phi$} &  27.11 & 26.90 & 23.22 & 23.48 & 19.65 & 20.00 & 16.66 & 16.90 \\
	\hline
	
	\makecell{Euclidean \\ with learned $\Phi$}	& 26.59 & 26.44 & 25.51 & 25.73 & 23.14 & 23.51 & 19.44 & 19.74 \\
			\hline	
		
	\makecell{Euclidean + Adversarial \\ with learned $\Phi$} & 30.53 & 29.42 & 26.47 & 25.94 & 22.98 & 23.00 & 19.06 & 19.31 \\
	\hline
	\end{tabular}
	\caption{This table shows the mean reconstruction PSNR on the test set for different variations of ReconNet i.e., with different loss functions and measurement matrices ($\Phi$). We see that the PSNR improves significantly at all measurement rates when a the measurement matrix is changed from a Gaussian matrix to a jointly learned one (Section \ref{sec:learn_mm}). We also observe that at higher measurement rates of 0.25 and 0.10, using adding adversarial loss to Euclidean loss (\ref{eq:advloss}) while training improves PSNR by about 1 dB in the case of a Gaussian $\Phi$ and about 3 dB when $\Phi$ is learned. }
	\label{table:gan}
\end{table*}


\section{Reconstruction of Real Data From Compressive Imager}
\label{sec:real_data}
The previous section demonstrated the superiority of our algorithm over traditional algorithms for simulated CS measurements. Here, we show that our networks trained on simulated data can be readily applied for real world scenario by reconstructing images from CS measurements obtained from our block SPC. We compare our reconstruction results with other algorithms.

\subsection{{\bf Scalable Optical Compressive Imager Testbed}}
\begin{figure}[]
	\centering
	\includegraphics[width=0.45\textwidth]{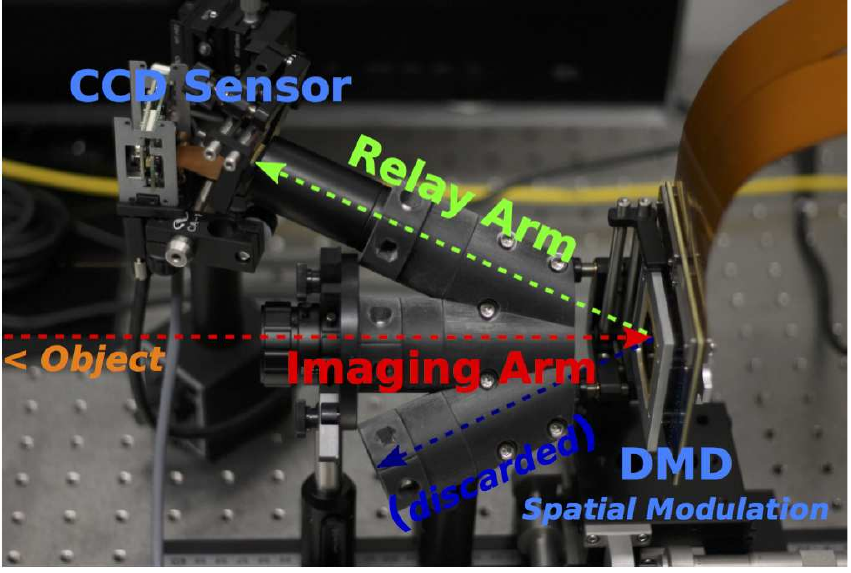}
	\caption{Compressive imager testbed layout with the object imaging arm in the center, the two DMD imaging arms are on the sides.}
	\label{fig:compressive_imager}
\end{figure}

Here we employ a compressive imaging system implementation \cite{SCIOSA15},\cite{kerviche2014information}, which is scalable with respect to field of view and/or resolution and avoids limitations inherent in a single-pixel implementation \cite{SPC}. Scalability is achieved via a block wise measurement  approach. The compressive imaging system is implemented via two imaging arms and a discrete mirror device (DMD) as shown in Figure \ref{fig:compressive_imager}. The DMD is an array of electronically controllable bi-stable mirrors of $10.8 \mu$m pitch, which modulates the incoming light intensity field with 8 bits gray-scale transmission patterns. The first arm of the system images the object or scene onto the DMD surface, mapping in to an area of about $262 \times 262$ micro-mirror element (or about $2.85$mm). The second images the DMD plane onto a detector array, which is 1/3" $640 \times 480$ CCD with a pixel pitch/size of $7.4 \mu$m operating at 12-bit quantization. Given the object, the DMD and the sensor planes that optical conjugates, the block of modulated patterns are each mapped to a small number of contiguous detectors whose outputs are digitally combined to return a single measurement per block. Thus the blocks and their mapping to group of detectors essentially behave like parallel Single Pixel Cameras (SPC). In this architecture, the modulation patterns on the DMD are generated by unfolding each row of the projection matrix $\Phi$, which are temporally scanned to acquire all the measurements, in parallel for all blocks.

It is important to highlight that one of the underlying challenges of implementing such a compressive imaging hardware is to ensure the correct calibration of the system, i.e. to minimize the deviation from the actual physical system measurement model to the idealized (and usually simplified) one. With this testbed, the we have partly automated this arduous calibration process. We employ uniform white object and display a series of known transmission patterns on the DMD to localize and identify the pixels of the sensor associated with a particular block. We refer to \cite{SCIOSA15} and \cite{kerviche2014information} for more details about this calibration process. This calibration process thus dynamically discovers the distorted mapping between the DMD plane and the sensor plane and also measures the bias and scaling non-uniformities across the blocks. Finally, the target images are shown on a display facing the imaging arm and the system, which are pre-corrected for the gamma correction applied by the display panel.

\subsection{{\bf Reconstruction experiments}} 
\begin{figure*}[ht!]
	\centering
	\begin{subfigure}[]{\textwidth}
	\includegraphics[trim = {2cm, 6cm, 3cm, 4cm}, clip,width=\textwidth]{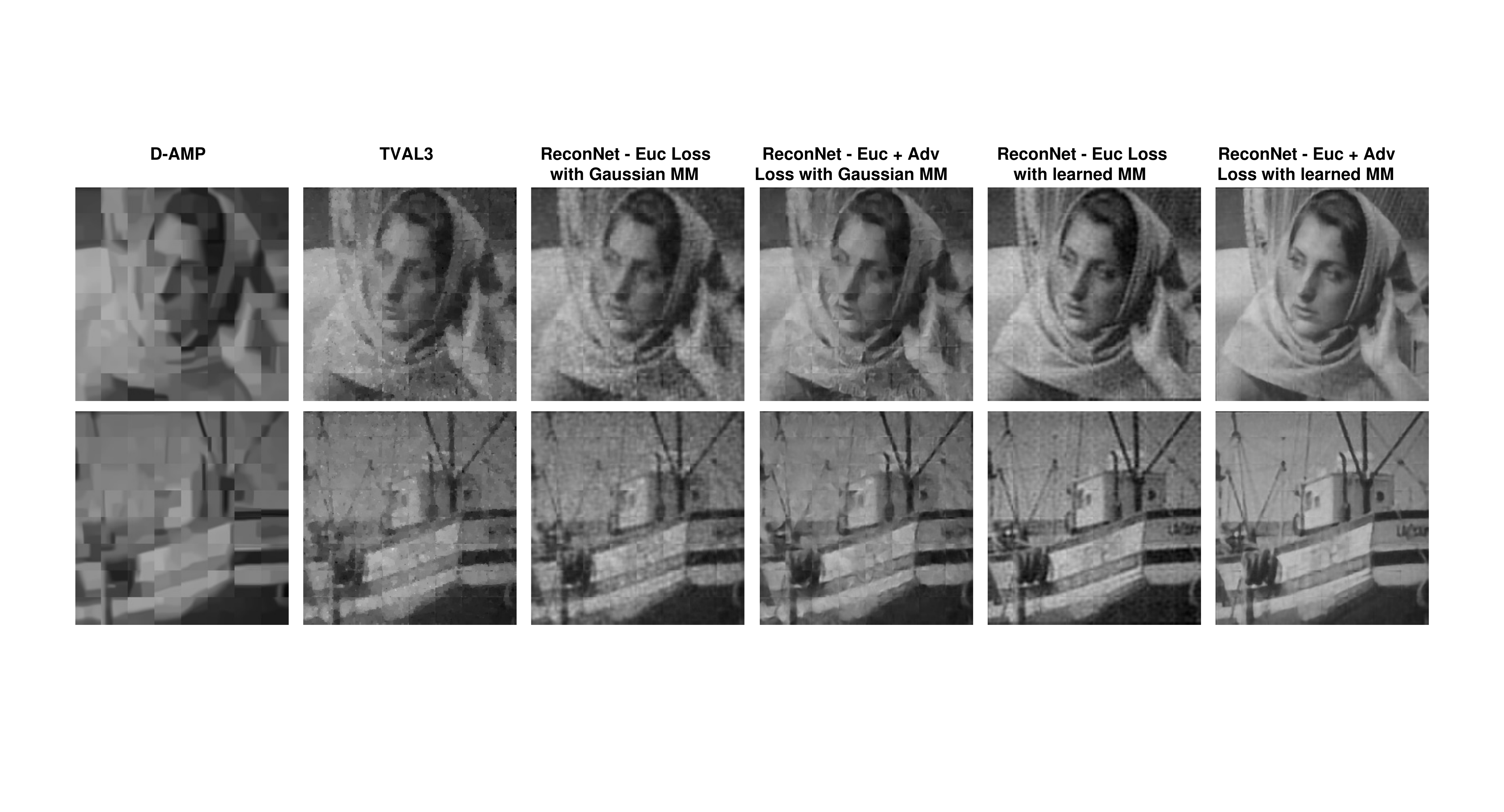}
	\caption{MR = 0.10}
	\label{fig:real_data_0_10}
	\end{subfigure}

\begin{subfigure}[]{\textwidth}
	\centering
	\includegraphics[trim = {2cm, 6cm, 3cm, 4cm}, clip, width=\textwidth]{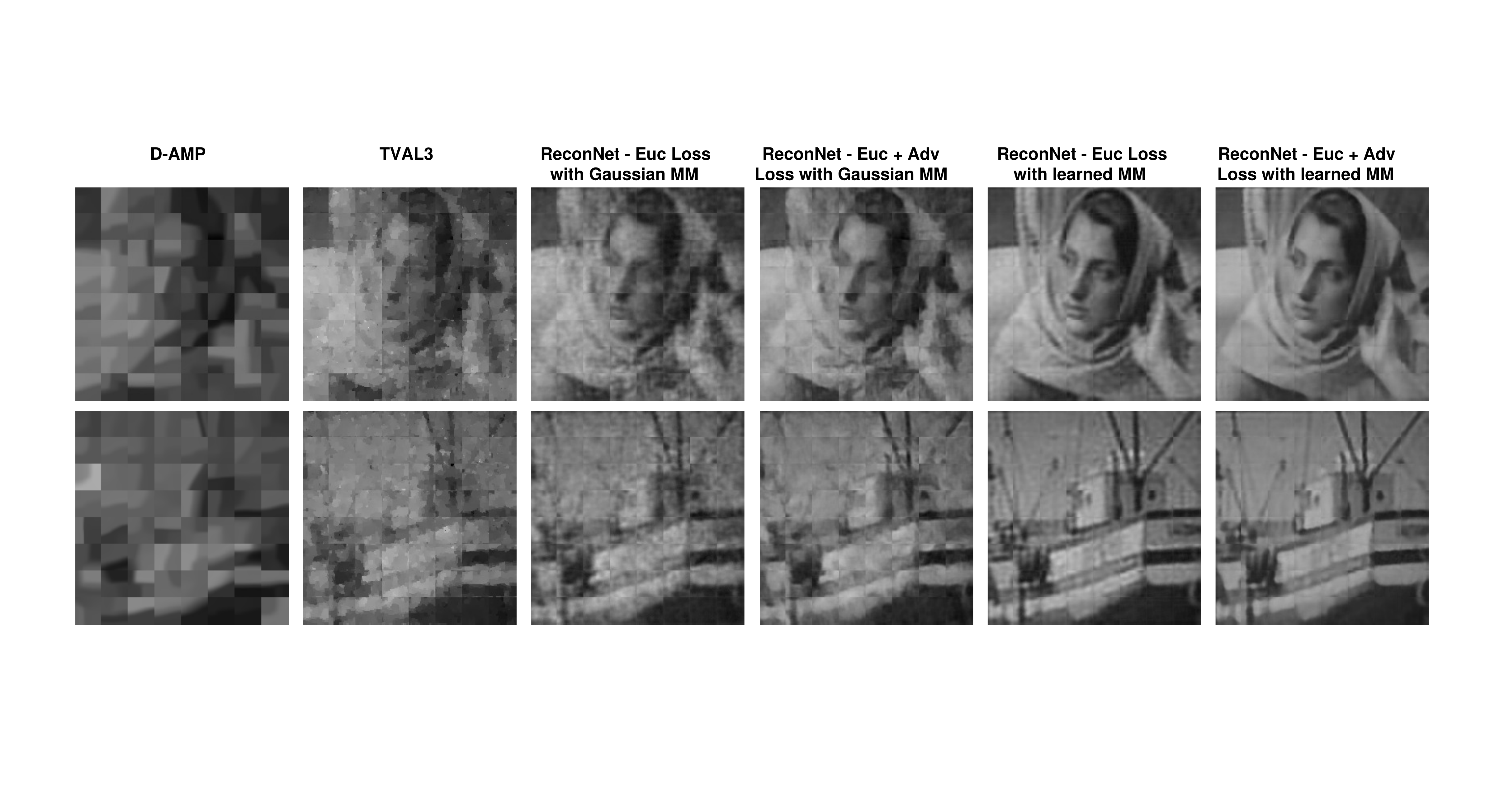}
	\caption{MR = 0.04}
		\label{fig:real_data_0_04}
\end{subfigure}
\caption{The figure shows reconstruction results for 2 images whose measurements are collected using our block SPC. The results are for two measurement rates (a) 0.10 and (b) 0.04. The iterative methods in first and second columns use a Gaussian $\Phi$. The next four columns shows the reconstructions obtained using different variants of ReconNet based on the loss function used in training and whether or not $\Phi$ was learned. ``MM" and $\Phi$ both stand for measurement matrix, ``Euc Loss" stands for Euclidean loss and ``Adv Loss" stands for adversarial loss. Clearly, all variants of ReconNet perform better than both TVAL3 and D-AMP. Both learning the measurement matrix and using adversarial loss tend to make reconstructions sharper and less noisy.}
\label{fig:real_data}
\end{figure*}

We use the set up described above to obtain CS measurements for each of the blocks (of size $33 \times 33$) in the scene. Operating at MR's of 0.1 and 0.04, we implement the 8-bit quantized versions of two kinds of measurement matrices: \begin{enumerate}
\item Orthogonalized random Gaussian matrices used to train networks in Section \ref{sec:training} and Section \ref{sec:gan}
\item Learned measurement matrices, from Section \ref{sec:learn_mm} and Section \ref{sec:gan}. In this case, the measurement matrices are implemented by the camera hardware by programming the DMD and the outputs are the CS measurements that are fed into the \textbf{second layer} of the trained networks directly.
\end{enumerate}
It is to be noted that the CS measurements are input to the  corresponding networks \textbf{trained on the simulated CS measurements}; no further training is done on the real data. Using these measurements we test four variants of ReconNet -- two kinds of measurement matrices (Gaussian or learned) and two kinds of loss functions(Euclidean or Euclidean + Adversarial Loss). Figures \ref{fig:real_data_0_10} and \ref{fig:real_data_0_04} show the reconstruction results MR = $0.10$ and $0.04$ respectively. The first and second columns show the reconstructions obtained using D-AMP and TVAL3 which are iterative algorithms. The next four columns show the results obtained using the four variants of ReconNet. It can be observed that our algorithm (all four variants) yields reconstructions that preserve more detail compared to the iterative approaches, thus demonstrating that our algorithm is robust to unseen sensor noise. For ReconNet, learning the measurement matrix improves results significantly. Using adversarial loss in addition to Euclidean loss while training yields sharper results as in the case of simulated CS data. Also, the degradation in reconstruction quality when measurement rate is reduced is less in the case of ReconNet than the iterative algorithms.

\section{Reducing Memory Footprint With Circulant Layers} \label{sec:circulant}
A drawback of the architecture presented in Section \ref{sec:architecture} is the large size of the first fully connected (FC) layer that maps the CS measurements to a 2D array. As a numerical example, consider ReconNet operating at an MR = 0.1 with a block size of $33 \times 33$. Then, the FC layer contains $109 * 1089 = 118701$ weights. By comparison, the rest of the layers are all convolutional and contain a total of 22720 parameters. In this section, we discuss ways to reduce the complexity of this layer.

In inference applications using deep learning such as image recognition, CNN architectures usually employ one or two fully connected layers at the end to map the convolutional feature maps to probability distributions over the class labels. Depending on the size of the feature maps, the number of classes etc., these FC layers tend to be large (relative to the rest of the network). Recent research has shown that we can reduce the complexity of these layers from $O(d^2)$ to $O(d)$ without any loss in performance. One particular paper is that of Cheng et al \cite{cheng2015exploration} which replaces the fully connected layer -- represented by a weight matrix without any constraints on the weights -- with a circulant layer where the weight matrix is constrained to be circulant matrix. They proceed to show that in spite of a large reduction in the number of parameters, the performance of the network largely remains the same and in some cases, even performs better! They also discuss how to efficiently compute the output of such a layer using FFTs. In this paper, we propose this layer as an alternative for the first fully connected layer.

A circulant matrix $C \in \mathbb{R}^{d \times d}$ is completely defined by a vector $\mathbf{c} = (c_0, c_1, ..., c_{d-1})$ as follows:
\vspace{-0.2cm}
\begin{equation}
\mathbf{C} = circ(\mathbf{c}) = \begin{bmatrix}
c_0 & c_{d-1} & \dots & c_2 & c_1 \\
c_1 & c_1 & \dots & c_3 & c_2 \\
\vdots & \vdots & \ddots & \vdots & \vdots \\
c_{d-1} & c_{d-2} & \dots & c_1 & c_0 
\end{bmatrix}.
\end{equation}

It can be shown that for an input $\mathbf{x} \in \mathbb{R}^d$, the output $\mathbf{y} \in \mathbb{R}^d$ of a circulant layer can be computed efficiently using

\begin{equation}
\mathbf{y} = \mathbf{C}\mathbf{x} = \mathbf{c} \circledast \mathbf{x} = \mathcal{F}^{-1}(\mathcal{F}(\mathbf{c}) \circ \mathcal{F}(\mathbf{x})), 
\end{equation}
where $\circledast$ represents circular convolution and $\circ$ is the element-wise multiplication operator. $\mathcal{F}$ and $\mathcal{F}^{-1}$ represent Fourier and inverse Fourier transforms respectively. We have implemented this layer in TensorFlow \cite{tensorflow2015-whitepaper} which computes the gradients using automatic differentiation.

In our case, the input vector $\mathbf{x} \in \mathbb{R}^M$ has a dimension less than that of the output of the first layer which is a vector with dimension equal to the number of pixels ($N$) in the block. Thus, in order to use the circulant layer instead of an FC layer, we will append $N - M$ zeros to each input $\mathbf{x}$ and hence, $\mathbf{C} \in \mathbb{R}^{N \times N}$ and $\mathbf{c} \in \mathbb{R^N}$. Therefore, the number of weights in the first layer of ReconNet can be reduced from $MN$ to $N$ by employing a circulant layer instead of an FC layer. For MR = 0.10, this corresponds to a 99.1\% reduction in parameters for the first layer. 

However, for higher measurement rates, this leads to significant under-fitting since the number of trainable parameters becomes small. We observed empirically that we can increase the reconstruction quality by using \textbf{multiple} circulant layers as the first layer instead of just one. At the output of the first layer, we have multiple feature maps from the circulant layers which are combined into a single tensor.  Thus, the convolutional layer that follows this layer must be modified. If the number of circulant layers is $\gamma$, then each filter in the following convolutional layer are of size $11 \times 11 \times \gamma$. This is only a modest increase in parameters for this layer compared to the $11 \times 11 \times 1$ filters which we would need in the case of a fully connected layer or a single circulant layer. 
 
We evaluate this by training networks at four measurement rates using ReconNet (Euc) as the network architecture. We increased the number of circulant layers from $1$ to a value $\gamma$ such that the reduction in the parameters of the first layer is no less than 95\% when compared to using a fully connected layer at the same measurement rate. The training and testing sets are same as in the previous sections. Table \ref{table:circulant} shows the mean PSNR obtained for the test set using ReconNet using circulant layer instead of an FC layer. We observe that the reduction in PSNR is within 2 dB at most measurement rates even with 95\% reduction in parameters of the first layer. 

\begin{table*}
	\footnotesize
	\centering
	\begin{tabular}{|c|c|c|c|c|c|c|}
		\hline
		\multirow{2}{*}{MR} & \multirow{2}{*}{\makecell{No. of\\Circulant Layers}} &  \multirow{2}{*}{\makecell{\% Reduction in Parameters \\ in the First Layer}} & \multicolumn{2}{c|}{Mean PSNR using circulant layers} & \multicolumn{2}{c|}{\makecell{Mean PSNR using an FC layer}} \\
		\cline{4-7}
		
		& & & without BM3D & with BM3D & without BM3D & with BM3D\\
		
		\hline
		\multirow{2}{*}{0.25} & 1 & 99.63 & 20.92 & 21.31 & \multirow{2}{*}{25.54} & \multirow{2}{*}{25.92}\\
		& 13 & 95.22 & 23.52 & 23.89 & & \\
		\hline
		
		\multirow{2}{*}{0.10} & 1 & 99.08 & 20.3 & 20.71 & \multirow{2}{*}{22.68} & \multirow{2}{*}{23.23} \\
		& 5 & 95.41 & 21.24 & 21.65 & & \\
		\hline
		
		\multirow{2}{*}{0.04} & 1 & 97.67 & 18.83 & 19.18 & \multirow{2}{*}{19.99} & \multirow{2}{*}{20.44}\\
		& 2 & 95.34 & 19.11 & 19.48 & & \\
		\hline
			
		0.01 & 1 & 90 & 16.51 & 16.77 & 17.27 & 17.55\\	
		\hline
	\end{tabular}
	\caption{Comparison of mean PSNR (in dB) of reconstruction of the test set using a one or more circulant layers instead of a fully connected layer as the first layer of ReconNet. We see that the reduction in PSNR using circulant layers is within 2 dB even with 95\% reduction in parameters in the first layer. (The entries in the last two columns are from Table \ref{table:psnr_test})}
	\label{table:circulant}
\end{table*}

\section{Real-time High Level Vision Using Compressive Imagers} \label{sec:tracking}

It is now clear that our CS reconstruction algorithm is non-iterative, real-time and capable of producing good quality reconstruction results, over a broad range of measurement rates. In this section, we demonstrate that despite the expected degradation in PSNR as the measurement rate is decreased to an extremely low value of 0.01 (10 measurements for a $33 \times 33$ block), our algorithm still yields reconstructions where rich semantic content is still retained. As stated earlier, in many resource-constrained inference applications the goal is to acquire the least amount of data required to perform effective high-level image understanding. 

\begin{figure*}
\begin{minipage}[c][8cm][t]{.5\textwidth}
  \vspace*{\fill}
  \centering
  \includegraphics[trim = {3cm, 0cm, 3cm, 0cm}, clip,width=\textwidth]{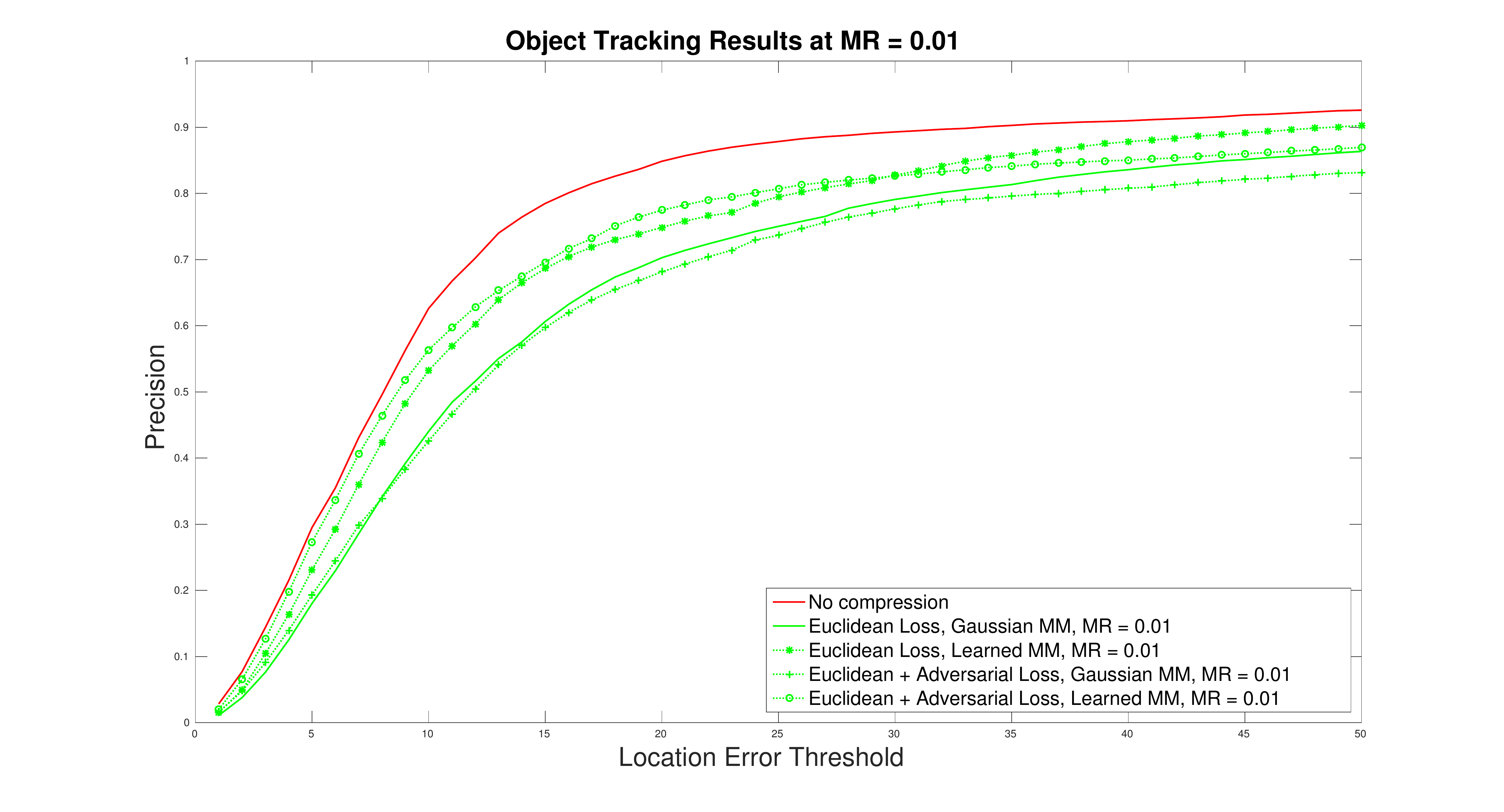}
  \subcaption{MR = 0.01}
\end{minipage}%
\begin{minipage}[c][8cm][t]{.5\textwidth}
\vspace*{\fill}
\centering
\includegraphics[trim = {3cm, 0cm, 3cm, 0cm}, clip,width=\textwidth]{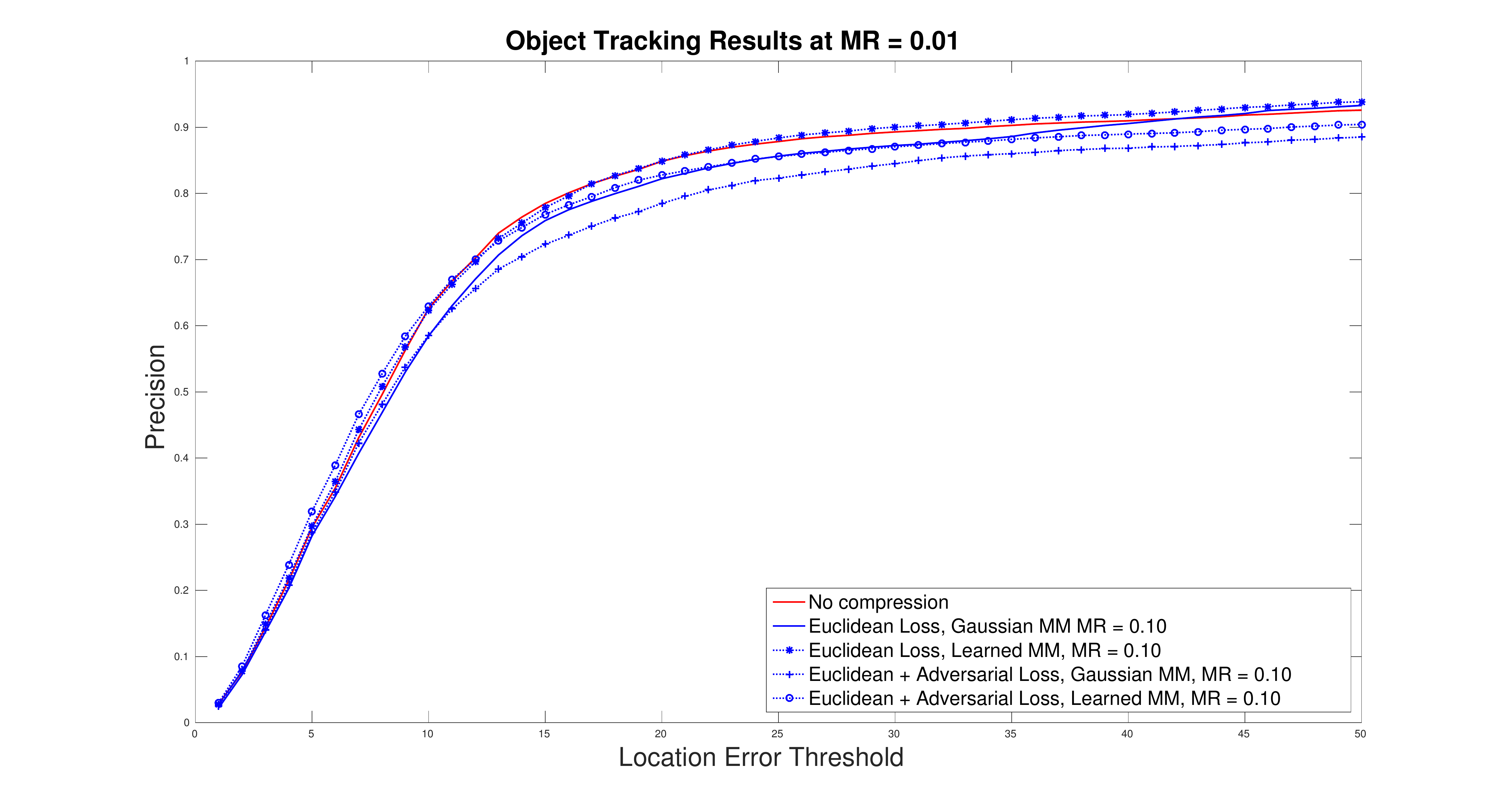}
\subcaption{MR = 0.10}
\end{minipage}%

\caption{The figure shows the variation of average precision with location error threshold for ReconNet+KCF and original videos. Clearly, semantic content required for object tracking is retained even in reconstructions at MR = 0.01}
\label{fig:tracking}
\end{figure*}

To demonstrate how CS imaging can applied in such scenarios, we present an example proof of concept real-time high level vision application - object tracking. To this end, we simulate frame-wise video compressive imaging at measurement rates of 0.01 and 0.10 by obtaining block CS measurements of each frame on 15 publicly available videos \cite{wu2015object} (BlurBody, BlurCar1, BlurCar2, BlurCar4, BlurFace, BlurOwl, Car2, CarDark, Dancer, Dancer2, Dudek, FaceOcc1, FaceOcc2, FleetFace, Girl2) used to benchmark tracking algorithms. Then, we perform object tracking on-the-fly as we recover the frames of the video using all the variants of ReconNet without the denoiser. For object tracking we use a state-of-the-art algorithm based on kernelized correlation filters \cite{henriques2015high}. We call this pipeline, ReconNet+KCF. For comparison, we conduct tracking on original videos as well. We use the default values of the tracking algorithm in all cases. Figure \ref{fig:tracking} shows the average precision curve over the 15 videos, in which each datapoint indicates the mean percentage of frames that are tracked correctly for a given location error threshold. Using a location error threshold of 20 pixels, the average precision over 15 videos for variants of ReconNet+KCF at MR = 0.01 is between 68.14\% and 77.46\%. At MR = 0.10, we obtain impressive tracking performance between 79.49\% and 84.89 \% for different variants. By comparison, tracking on the original videos yields an average precision value of 84.9\%. Learning the measurement matrix gives a significant boost of about 8 and 5 percentage points at MR = 0.01 and 0.10 respectively. 

The effect of loss function is more nuanced. Euclidean + Adversarial loss seems to decrease tracking performance at MR = 0.10 with any measurement matrix and at MR = 0.01 with a Gaussian measurement matrix by about 3 percentage points over a large range of location error thresholds when compared to just Euclidean loss. However, we observe the opposite in the case of MR = 0.01 using a learned measurement matrix. Here, Euclidean + Adversarial loss outperforms Euclidean loss by about 3 percentage points. ReconNet + KCF operates at around 10 Frames per Second (FPS) for a video with frame size of $480 \times 720$ to as high as 56 FPS for a frame size of $240 \times 320$. 

\section{Conclusion}
In this paper we have described ReconNet -- a non-iterative algorithm for CS image reconstruction based on CNNs. The advantages of this algorithm are two-fold -- it can be easily implemented while making it 3 orders of magnitude faster than traditional iterative algorithms essentially making reconstruction real-time and it provides excellent reconstruction quality retaining rich semantic information over a large range of measurement rates. We have also discussed novel ways to improve the basic version of our algorithm. We have proposed learning the measurement matrix jointly with the reconstruction network as well as training with adversarial loss based on recently popular GANs. In both cases, we have shown significant  improvements in reconstruction quality over a range of measurement rates. Using the ReconNet + KCF pipeline, efficient real-time tracking is possible using CS measurements even at a very low measurement rate of 0.01. This also means that other high-level inference applications such as image recognition can be performed using a similar framework i.e., ReconNet + Recognition from CS measurements. We hope that this work will generate more interest in building practical real-world devices and applications for compressive imaging. 

\appendices
\section{Additional Results}
\begin{figure*}
\begin{minipage}[c][10cm][t]{0.45\textwidth}
  \vspace*{\fill}
  \centering
  \includegraphics[trim = {3cm, 3cm, 5cm, 1cm}, clip,width=\textwidth]{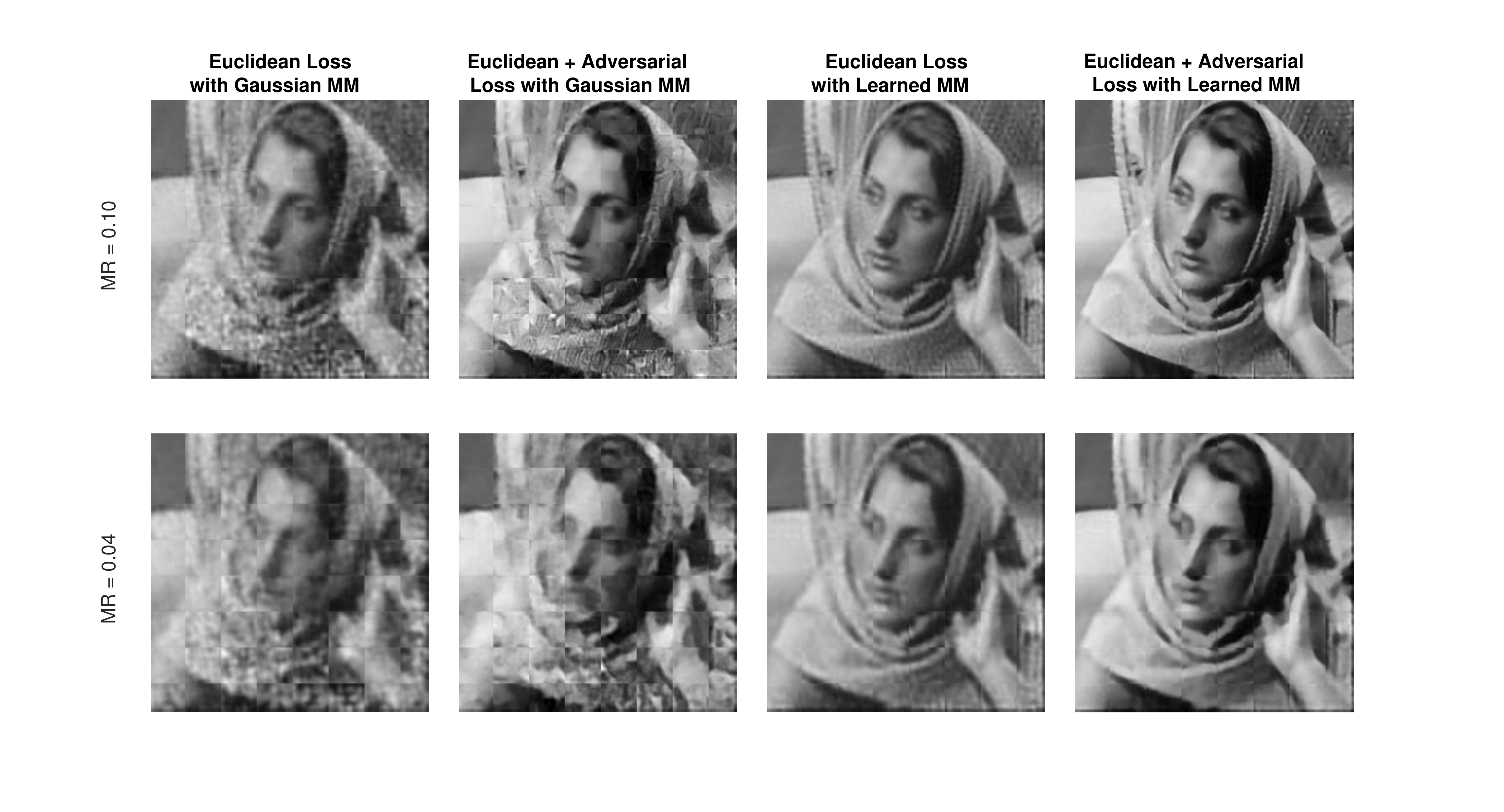}
  \subcaption{Barbara}
  \par\vfill
  \includegraphics[trim = {3cm, 3cm, 5cm, 1cm}, clip,width=\textwidth]{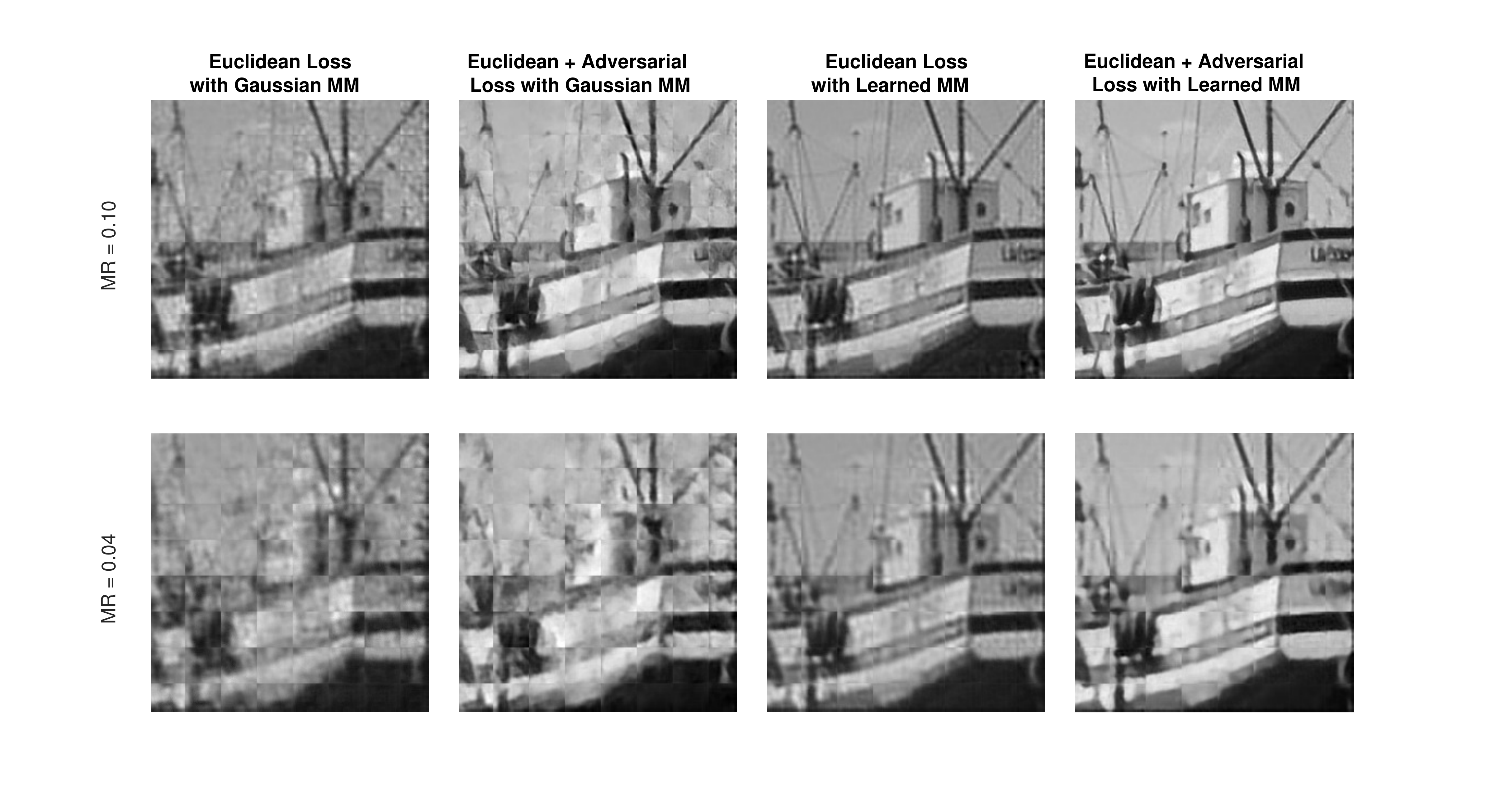}
  \subcaption{Boats}
  \par\vfill
  \includegraphics[trim = {3cm, 3cm, 5cm, 1cm}, clip,width=\textwidth]{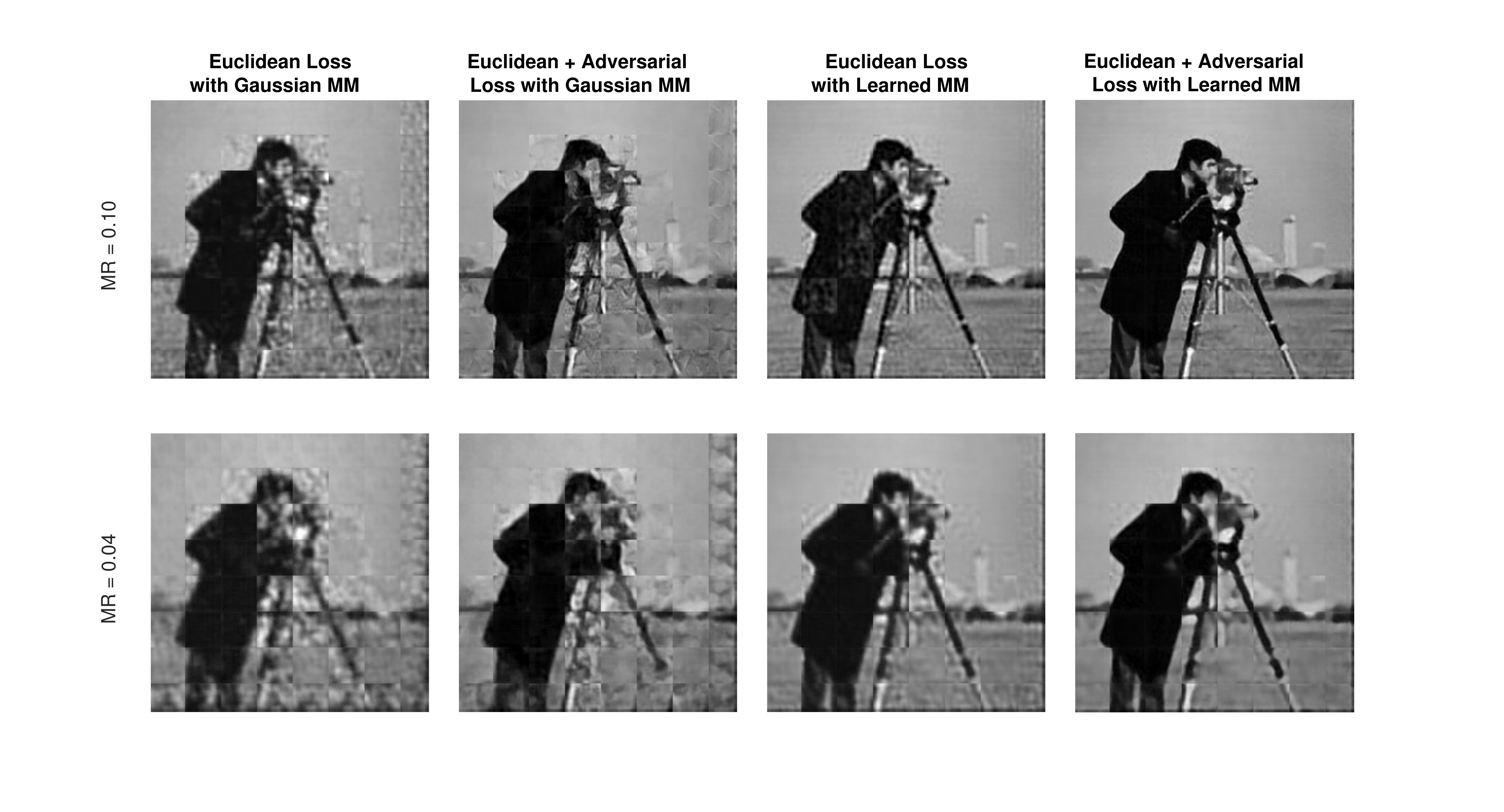}
  \subcaption{Cameraman}
  \par\vfill
  \includegraphics[trim = {3cm, 3cm, 5cm, 1cm}, clip,width=\textwidth]{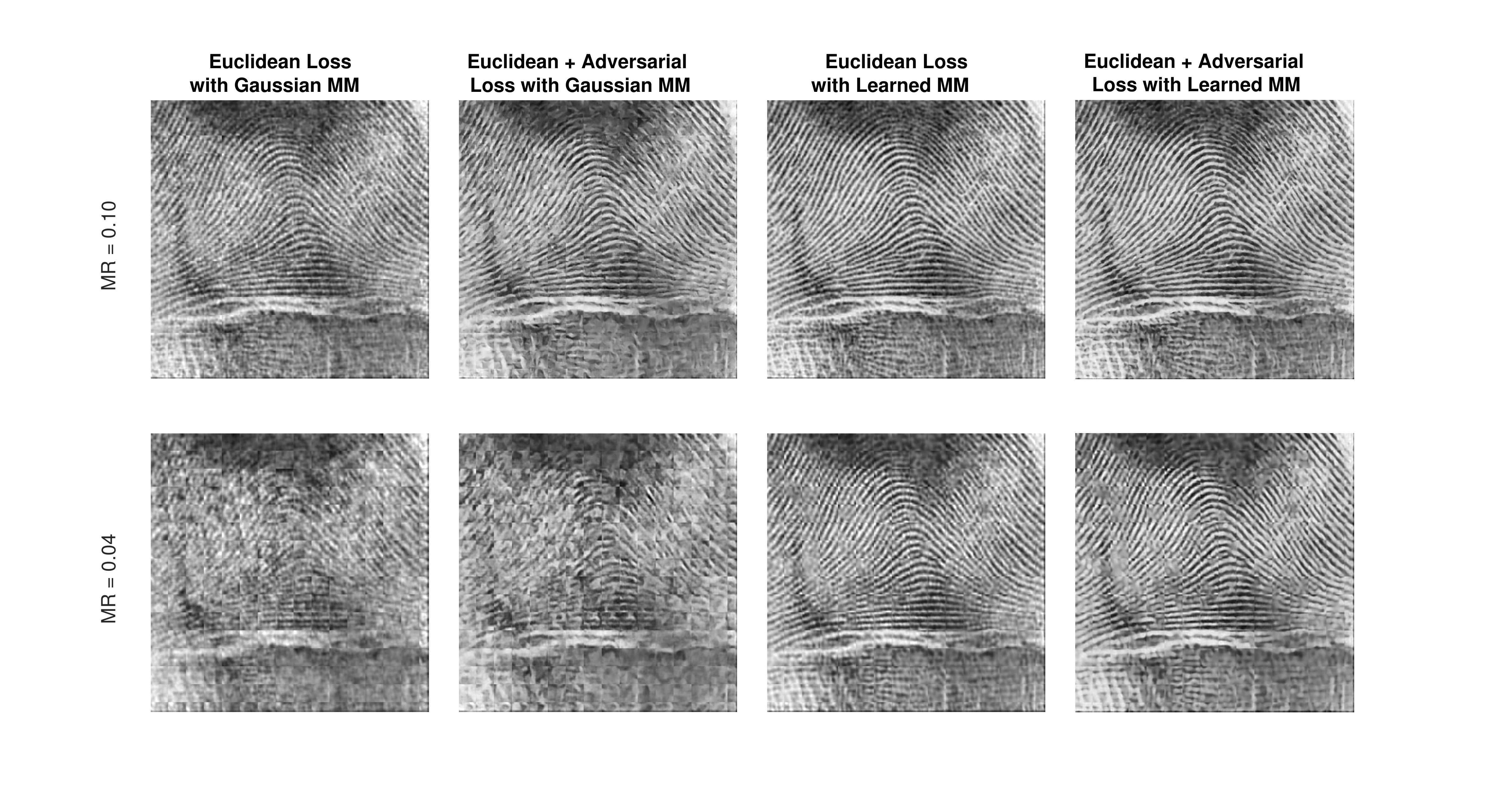}
  \subcaption{Fingerprint}
\end{minipage}%
\begin{minipage}[c][10cm][t]{.45\textwidth}
  \vspace*{\fill}
  \centering
  \includegraphics[trim = {3cm, 3cm, 5cm, 1cm}, clip,width=\textwidth]{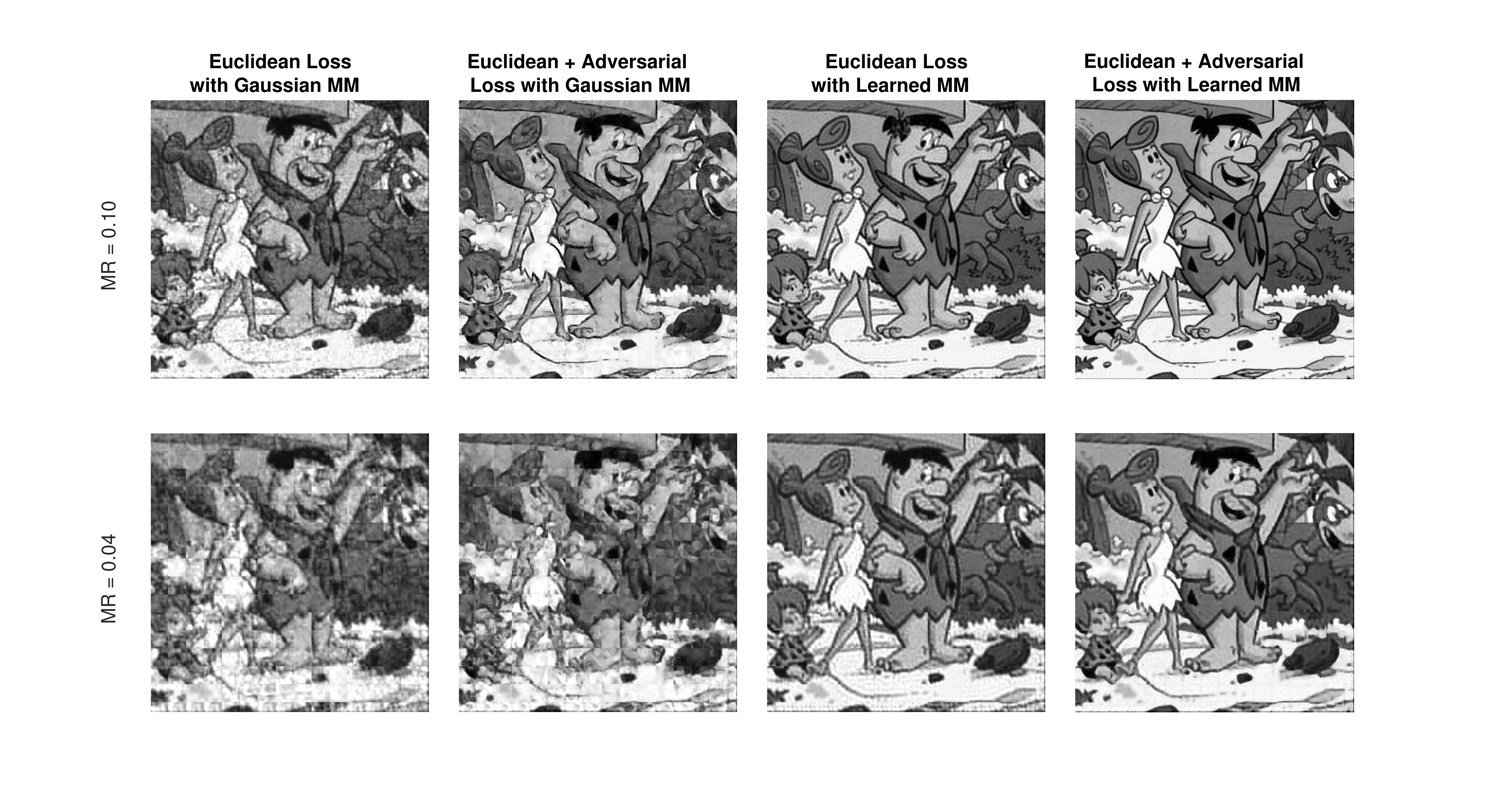}
  \subcaption{Flintstones}
  \par\vfill
  \includegraphics[trim = {3cm, 3cm, 5cm, 1cm}, clip,width=\textwidth]{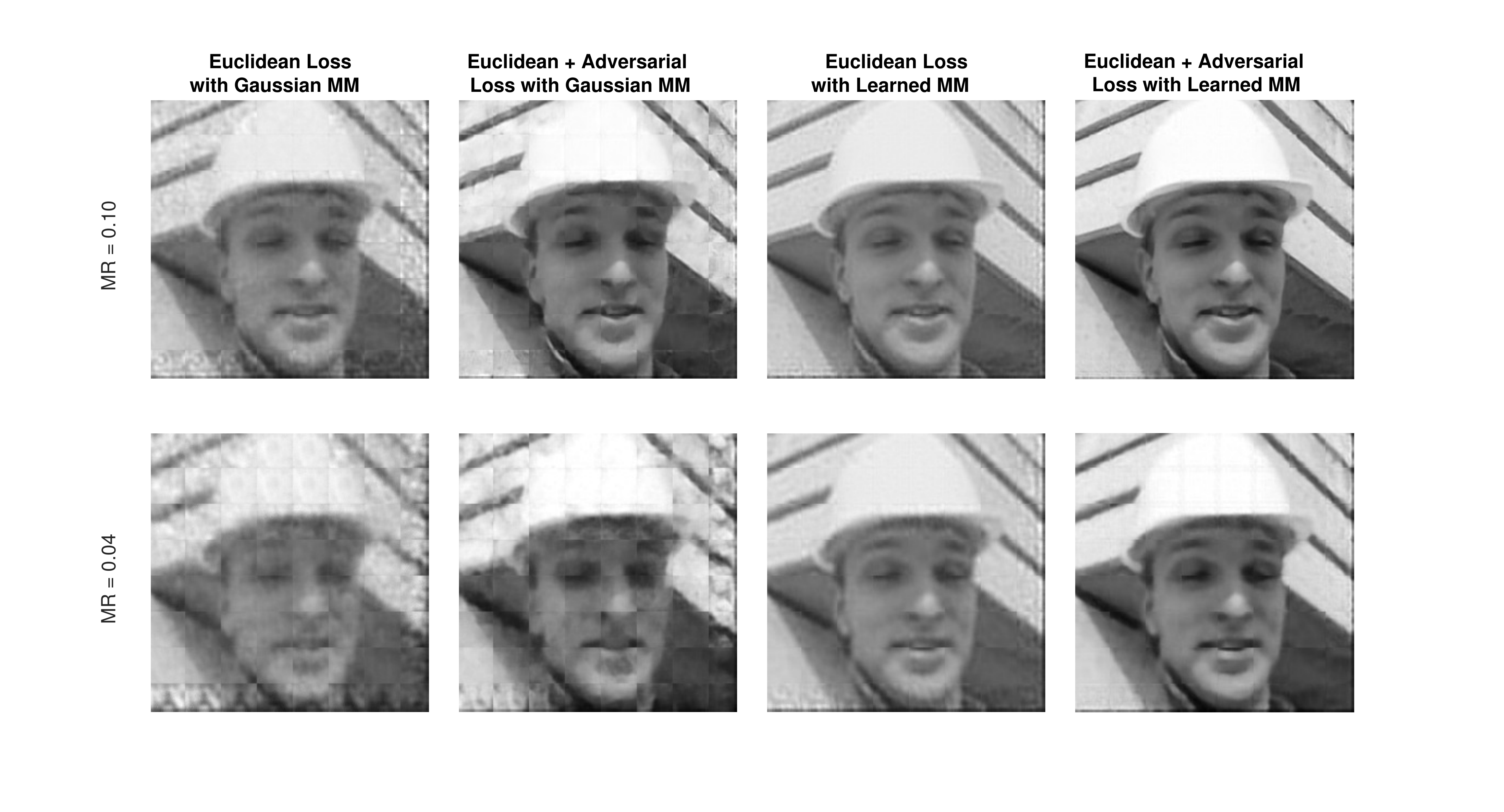}
  \subcaption{Foreman}
  \par\vfill
  \includegraphics[trim = {3cm, 3cm, 5cm, 1cm}, clip,width=\textwidth]{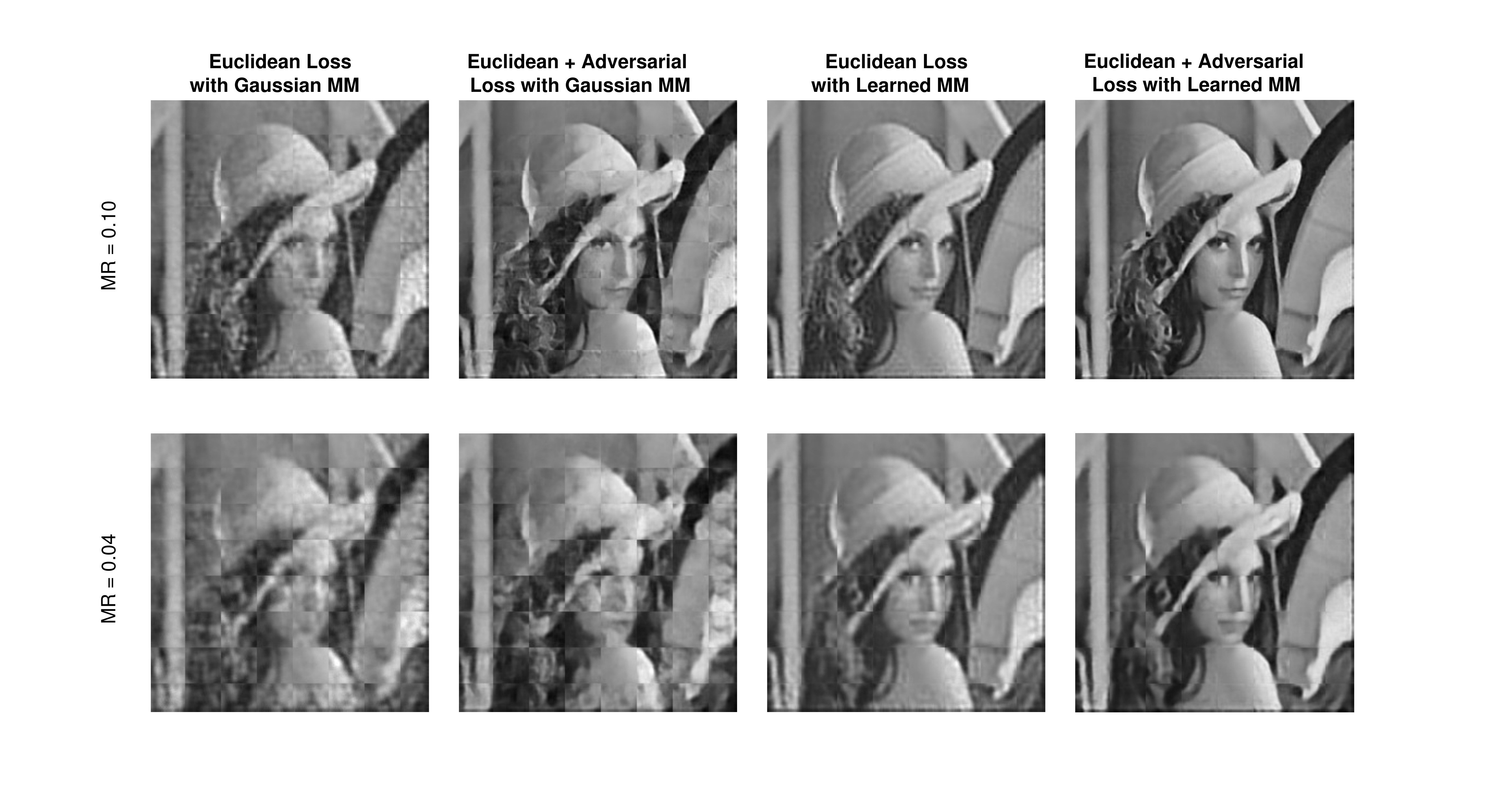}
  \subcaption{Lena}
  \par\vfill
  \includegraphics[trim = {3cm, 3cm, 5cm, 1cm}, clip,width=\textwidth]{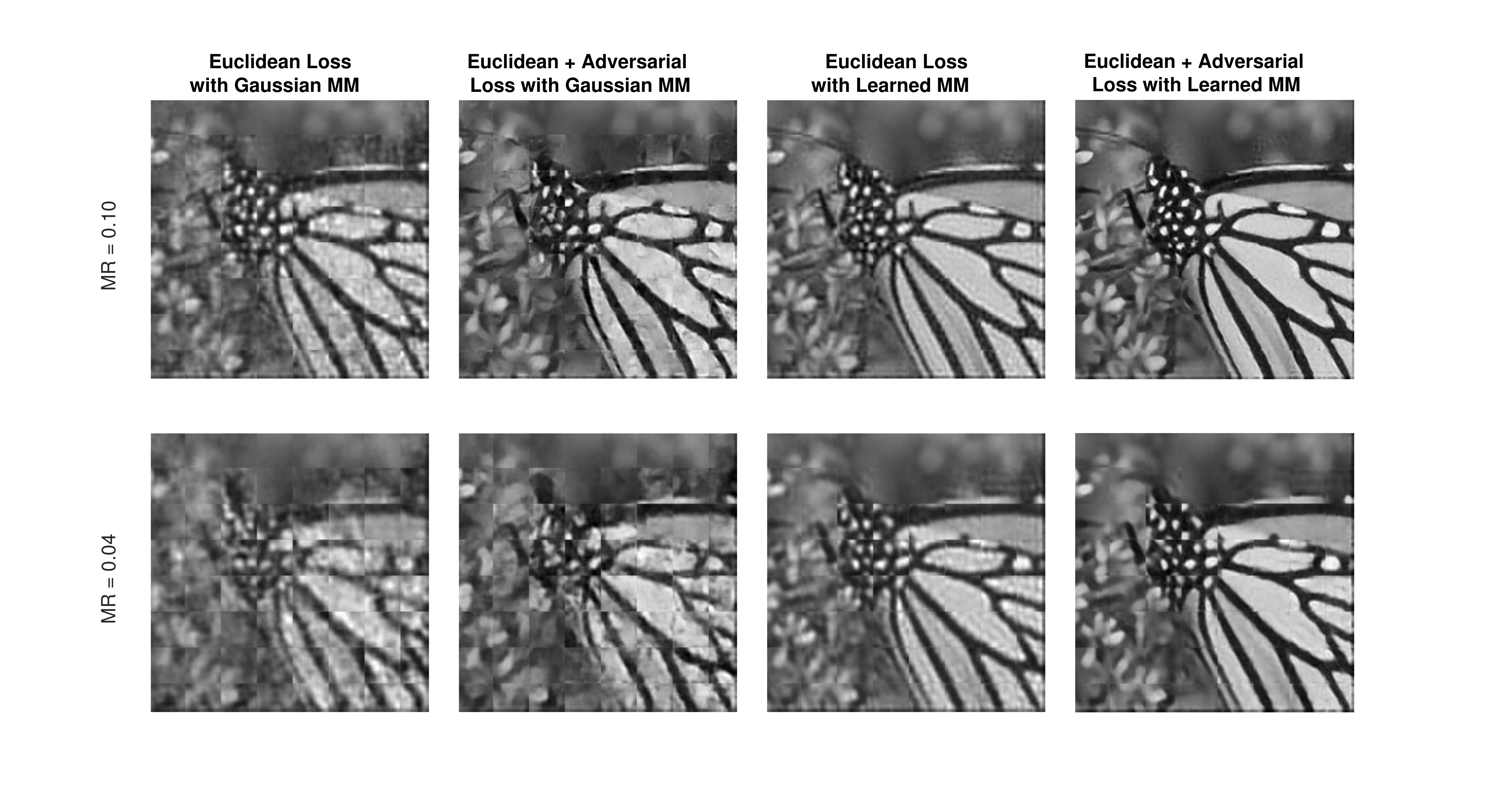}
  \subcaption{Monarch}
\end{minipage}%
\vspace{4.5in}
\caption{The figures show reconstruction results for the some test images at two measurement rates of 0.1 and 0.04 from measurements obtained using different variants of ReconNet. We can clearly observe that learning the measurement matrix as well as using adversarial loss while training produce superior quality reconstruction (both independently and together) at both measurement rates when compared to the basic version of ReconNet. MM refers to the measurement matrix.}
\label{fig:stuff}
\end{figure*}

In Table \ref{table:psnr_test}, we presented the peak signal to-noise ratio (PSNR) values for 4 of the 11 test images. Here, the PSNR values (in dB) for the remaining 7 test images for various measurement rates are presented in Table \ref{table:psnr_test_remaining}. In Figure \ref{fig:learn_vs_gaussian}, reconstructions using the 4 variants of our algorithm for 2 test images were shown. The reconstructions of 8 additional images are shown in Figure \ref{fig:stuff}. 

\begin{table*}
	\footnotesize
	\centering
	\begin{tabular}{|c|c|c|c|c|c|c|c|c|c|}
		\hline
		\multirow{2}{*}{\makecell{Image \\ Name}} & 
		\multirow{2}{*}{Algorithm} & 
		\multicolumn{2}{c|}{MR = 0.25} & 
		\multicolumn{2}{c|}{MR = 0.10} & 
		\multicolumn{2}{c|}{MR = 0.04} & 
		\multicolumn{2}{c|}{MR = 0.01} \\
		\cline{3-10}
		& & w/o BM3D & w/ BM3D & w/o BM3D & w/ BM3D & w/o BM3D & w/ BM3D & w/o BM3D & w/ BM3D\\
		\hline

 		\multirow{5}{*}{Parrot} & TVAL3 \cite{li2013efficient} & \textbf{27.17} & \textbf{27.24} & 23.13 & \textbf{23.16} & 18.88 & 18.90 & 11.44 & 11.46\\
 		& NLR-CS \cite{dong2014compressive} & 26.53 & 26.72 & 14.14 & 14.16 & 10.59 & 10.92 & 5.11 & 5.44\\
 		& D-AMP \cite{metzler2014denoising} & 26.86 & 26.99 & 21.64 & 21.64 & 15.78 & 15.78 & 5.09 & 5.09\\
 		& SDA \cite{mousavi2015deep} & 24.48 & 24.36 & 22.13 & 22.35 & \textbf{20.37} & 20.67 & \textbf{17.70} & 17.88\\
 		& \makecell{\textbf{ReconNet} \\ (Euc) \cite{Kulkarni_2016_CVPR}} & 25.59 & 26.22 & 22.63 & \textbf{23.23} & 20.27 & \textbf{21.06} & 17.63 & \textbf{18.30}\\
 		& \makecell{\textbf{ReconNet} \\ (Euc + Adv)} & 27.14 & 27.63 & 23.52 & 23.94 & 20.67 & 21.11 & 17.69 & 17.97 \\

 		\hline
		
		\multirow{5}{*}{Barbara} & TVAL3 & 24.19 & 24.20 & 21.88 & 22.21 & 18.98 & 18.98 & 11.94 & 11.96\\
		& NLR-CS & \textbf{28.01} & \textbf{28.00} & 14.80 & 14.84 & 11.08 & 11.56 & 5.50 & 5.86\\
		& D-AMP & 25.89 & 25.96 & 21.23 & 21.23 & 16.37 & 16.37 & 5.48 & 5.48\\
		& SDA & 23.19 & 23.20 & 22.07 & 22.39 & \textbf{20.49} & 20.86 & 18.59 & 18.76\\
		& \makecell{\textbf{ReconNet} \\ (Euc)} & 23.25 & 23.52 & \textbf{21.89} & \textbf{22.50} & 20.38 & \textbf{21.02} & \textbf{18.61} & \textbf{19.08}\\
		& \makecell{\textbf{ReconNet} \\ (Euc + Adv)} & 23.78 & 23.46 & 20.91 & 21.00 & 19.00 & 19.50 & 16.91 & 17.23 \\
		\hline
		
 		\multirow{5}{*}{Boats} & TVAL3 & 28.81 & 28.81 & 23.86 & 23.86 & 19.20 & 19.20 & 11.86 & 11.88\\
 		& NLR-CS & 29.11 & \textbf{29.27} & 14.82 & 14.86 & 10.76 & 11.21 & 5.38 & 5.72\\
 		& D-AMP & \textbf{29.26} & 29.26 & 21.95 & 21.95 & 16.01 & 16.01 & 5.34 & 5.34\\
 		& SDA & 26.56 & 26.25 & 24.03 & \textbf{24.18} & 21.29 & 21.54 & \textbf{18.54} & 18.68\\
 		& \makecell{\textbf{ReconNet} \\ (Euc)} & 27.30 & 27.35 & \textbf{24.15} & 24.10 & \textbf{21.36} & \textbf{21.62} & 18.49 & \textbf{18.83}\\
 		& \makecell{\textbf{ReconNet} \\ (Euc + Adv)} & 27.72 & 26.93 & 23.68 & 23.60 & 19.84 & 20.18 & 16.80 & 17.02 \\

 		\hline

 		\multirow{5}{*}{Cameraman} & TVAL3 & \textbf{25.69} & \textbf{25.70} & \textbf{21.91} & \textbf{21.92} & 18.30 & 18.33 & 11.97 & 12.00\\
 		& NLR-CS & 24.88 & 24.96 & 14.18 & 14.22 & 11.04 & 11.43 & 5.98 & 6.31\\
 		& D-AMP & 24.41 & 24.54 & 20.35 & 20.35 & 15.11 & 15.11 & 5.64 & 5.64\\
 		& SDA & 22.77 & 22.64 & 21.15 & 21.30 & \textbf{19.32} & 19.55 & 17.06 & 17.19\\
 		& \makecell{\textbf{ReconNet} \\ (Euc)} & 23.15 & 23.59 & 21.28 & 21.66 & 19.26 & \textbf{19.72} & \textbf{17.11} & \textbf{17.49}\\
 		& \makecell{\textbf{ReconNet} \\ (Euc + Adv)} & 25.11 & 25.20 & 21.94 & 22.18 & 19.58 & 19.95 & 17.09 & 17.37 \\

		\hline
		
		\multirow{5}{*}{Foreman} & TVAL3 & 35.42 & 35.54 & \textbf{28.69} & \textbf{28.74} & 20.63 & 20.65 & 10.97 & 11.01\\
 		& NLR-CS & \textbf{35.73} & \textbf{35.90} & 13.54 & 13.56 & 9.06 & 9.44 & 3.91 & 4.25\\
 		& D-AMP & 35.45 & 34.04  & 25.51 & 25.58 & 16.27 & 16.78 & 3.84 & 3.83\\
 		& SDA & 28.39 & 28.89  & 26.43 & 27.16 & 23.62 & 24.09 & \textbf{20.07} & 20.23\\
 		& \makecell{\textbf{ReconNet} \\ (Euc)} & 29.47 & 30.78 & 27.09 & 28.59 & \textbf{23.72} & \textbf{24.60} & 20.04 & \textbf{20.33}\\
		& \makecell{\textbf{ReconNet} \\ (Euc + Adv)} & 31.26 & 32.17 & 27.42 & 28.31 & 23.09 & 23.76 & 18.74 & 19.08 \\
		
 		\hline

		\multirow{5}{*}{Lena} & TVAL3 & 28.67 & 28.71 & \textbf{24.16} & 24.18 & 19.46 & 19.47 & 11.87 & 11.89\\
		& NLR-CS & \textbf{29.39} & \textbf{29.67} & 15.30 & 15.33 & 11.61 & 11.99 & 5.95 & 6.27\\
		& D-AMP & 28.00 & 27.41 & 22.51 & 22.47 & 16.52 & 16.86 & 5.73 & 5.96\\
		& SDA & 25.89 &  25.70 & 23.81 & 24.15 & 21.18 & 21.55 & 17.84 & 17.95\\
		& \makecell{\textbf{ReconNet} \\ (Euc)} & 26.54 & 26.53 & 23.83 & \textbf{24.47} & \textbf{21.28} & \textbf{21.82} & \textbf{17.87} & \textbf{18.05}\\
		& \makecell{\textbf{ReconNet} \\ (Euc + Adv)} & 27.99 & 27.65 & 24.35 & 24.65 & 20.61 & 21.11 & 17.51 & 17.83 \\
		
 		\hline

 		\multirow{5}{*}{Peppers} & TVAL3 & 29.62 & 29.65 & \textbf{22.64} & 22.65 & 18.21 & 18.22 & 11.35 & 11.36\\
 		& NLR-CS & 28.89 & 29.25 & 14.93 & 14.99 & 11.39 & 11.80 & 5.77 & 6.10\\
 		& D-AMP & \textbf{29.84} & \textbf{28.58} & 21.39 & 21.37 & 16.13 & 16.46 & 5.79 & 5.85\\
 		& SDA & 24.30 & 24.22 & 22.09 & 22.34 & \textbf{19.63} & 19.89 & \textbf{16.93} & \textbf{17.02}\\
		& \makecell{\textbf{ReconNet} \\ (Euc)} & 24.77 & 25.16 & 22.15 & \textbf{22.67} & 19.56 & \textbf{20.00} & 16.82 & 16.96\\
		& \makecell{\textbf{ReconNet} \\ (Euc + Adv)} & 27.90 & 27.90 & 23.68 & 24.09 & 19.84 & 20.29 & 16.93 & 17.16 \\
		
		\hline

		\hline
	\end{tabular}
	\caption{{PSNR values in dB for 7 test images using different algorithms at different measurement rates. At low measurement rates of 0.1, 0.04 and 0.01, both variants of our algorithm yields superior quality reconstructions than the traditional iterative CS reconstruction algorithms, TVAL3, NLR-CS, and D-AMP.}}
	\label{table:psnr_test_remaining}
\end{table*}

\ifCLASSOPTIONcaptionsoff
  \newpage
\fi

\bibliographystyle{IEEEtran}

\bibliography{IEEEabrv,main}

\end{document}